\pdfoutput=1

\documentclass{article}
\usepackage{colm2024_conference}

\usepackage{url}
\usepackage{booktabs}
\usepackage{hyperref}
\usepackage{graphicx}
\usepackage{microtype}
\usepackage{multicol}
\usepackage{xcolor}   
\usepackage{multirow}
\usepackage{lipsum}
\usepackage{enumitem}
\usepackage{pifont}
\usepackage{colortbl}
\usepackage{tabularx}
\usepackage{caption} 
\usepackage{float}
\usepackage{amsmath,amsfonts,amssymb,bbm}
\usepackage{wrapfig}
\usepackage{caption}
\usepackage[T1]{fontenc}
\usepackage[most]{tcolorbox}

\tcbset{
  colback=cyan!3,
  colframe=cyan!40,
  coltitle=black,
  fonttitle=\bfseries,
  boxrule=0.6pt,             
  arc=2pt,                   
  left=6pt,
  right=6pt,
  top=6pt,
  bottom=6pt,
  enhanced,
  breakable                  
}

\usepackage{listings}

\lstdefinestyle{promptstyle}{
  basicstyle=\ttfamily\small,
  breaklines=true,
  breakatwhitespace=true,
  columns=fullflexible,
  keepspaces=true,
  showstringspaces=false,
  frame=none
}

\title{HLE-Verified: A Systematic Verification and Structured \\Revision of Humanity's Last Exam}

\author{
\vspace{1.5mm}
Weiqi Zhai$^{1*}$,
Zhihai Wang$^{2*}$,
Jinghang Wang$^{1}$,
Boyu Yang$^{1}$,
Xiaogang Li$^{1}$,
Xander Xu$^{1}$,
Bohan Wang$^{2}$,
Peng Wang$^{2}$,
Xingzhe Wu$^{2}$,
Anfeng Li$^{2}$,
Qiyuan Feng$^{1}$,
Yuhao Zhou$^{1}$,
Taolin Han$^{1}$,
Wenjie Luo$^{1}$,
Yiyuan Li$^{1}$,
Xiang Zheng$^{1}$,
Yaxuan Wang$^{1}$,
Ruixiang Luo$^{1}$,
Guojie Lin$^{1}$,
Peiyao Xiao$^{1}$,
Chengliang Xu$^{1}$,
Ben Wang$^{1}$,
Zeyu Wang$^{1}$,
Zichao Chen$^{1}$,
Jianan Ye$^{1}$,
Yijie Hu$^{1}$,
Jialong Chen$^{1}$,
Zongwen Shen$^{1}$,
Yuliang Xu$^{1}$,
An Yang$^{2}$,
Bowen Yu$^{2}$,
Dayiheng Liu$^{2}$,
Junyang Lin$^{2}$,
Hu Wei$^{1}$,
Que Shen$^{2\dagger}$,
Bing Zhao$^{1\dagger}$ \\
\vspace{2mm}
$^1$Alibaba Group \\
$^2$Qwen Team, Alibaba Group \\
\vspace{1mm}
\small
* Equal Contribution \\
$\dagger$ Corresponding Authors: shenque.sq@alibaba-inc.com, xiongdao@alibaba-inc.com
}

\begin{document}
\maketitle

\begin{abstract}
Humanity's Last Exam (HLE) has become a widely used benchmark for evaluating frontier large language models on challenging, multi-domain questions. However, subsequent community-led analyses have raised concerns that HLE contains a non-trivial number of noisy items (e.g., ambiguous statements, incorrect answers, or mismatched rationales), which can systematically bias evaluation results and distort cross-model comparisons. To address this challenge, we introduce \textbf{HLE-Verified}, a verified and revised version of HLE accompanied by a transparent, component-wise verification protocol and fine-grained error taxonomy. Our construction follows a two-stage validation-and-repair workflow resulting in a unified certified benchmark. In \textbf{Stage I}, each item is subjected to binary validation on the \emph{problem} and \emph{final answer} dimensions (with rationale used as an auxiliary consistency signal), combining domain-expert review and model-based cross-checks. This stage yields \textbf{668} items verified as correct. In \textbf{Stage II}, items identified as flawed but fixable are systematically revised under strict constraints that preserve the original evaluation intent, through dual independent expert repairs, model-assisted consistency auditing, and final expert adjudication, resulting in \textbf{1,143} revised-and-certified items. The remaining \textbf{689} items are released as a documented \emph{uncertain} set with explicit uncertainty sources and required expertise tags for future community refinement. We compare the performance of eight state-of-the-art language models on HLE and HLE-Verified, observing an average absolute accuracy gain of \textbf{7--10 percentage points} on HLE-Verified. The improvement is particularly pronounced on items where the original HLE problem statement and/or reference answer is erroneous: on this subset, the models achieve an average accuracy increase of \textbf{30--40 percentage points}. Moreover, our analyses indicate a strong association between model confidence and the presence of errors in the problem statement or reference answer, providing evidence for the effectiveness of our revisions. Overall, HLE-Verified improves HLE-style evaluations by reducing annotation noise and enabling more faithful measurements of model capabilities. Data is available at: https://huggingface.co/datasets/skylenage/HLE-Verified.
\end{abstract}

\section{Introduction}

As frontier language models (LMs) continue to advance rapidly, there is increasing demand for evaluation benchmarks that are simultaneously difficult, broad in disciplinary coverage, and resistant to saturation. 
High-difficulty benchmarks play a central role in tracking progress, validating capability claims, and shaping scientific discourse about model reasoning abilities. Against this backdrop, the Humanity's Last Exam (HLE) has emerged as a widely adopted benchmark for evaluating model performance on challenging questions across mathematics, science, engineering, and the humanities\citep{phan2025humanitysexam}. Results on HLE have been cited to support claims regarding reasoning depth, generalization capacity, and reliability of state-of-the-art models.

However, as benchmark difficulty increases, so does the importance of annotation integrity. When many items lie near a model’s decision boundary, even minor inconsistencies in problem statements or reference answers can disproportionately influence aggregate metrics. 
Recent independent audits and community-led reviews have highlighted substantial noise in the HLE dataset, including ambiguous or poorly defined question statements, erroneous reference answers, and inconsistencies between rationales and final answers \citep{white2025hle_wrong_answers,hle_gpqa_error_claims_github,reddit_forensic_audit_hle_2025}. These analyses suggest that a portion of measured model performance on HLE may reflect annotation artifacts rather than genuine capability differences.

This issue highlights a broader methodological question: how should high-difficulty benchmarks be systematically verified and maintained after deployment? While benchmark construction traditionally emphasizes task design and data collection, comparatively less attention has been devoted to post-release verification, structured auditing, and transparent revision protocols. For benchmarks that inform scientific claims and cross-model comparisons, the absence of systematic validation can undermine interpretability, reproducibility, and measurement reliability.

To address this reliability challenge, we developed \textbf{HLE-Verified}—a rigorously validated and partially revised version of HLE built under a structured two-stage validation and revision protocol. This approach employs a component-level verification framework, treating \emph{problem statements} and \emph{final answers} as primary targets for correctness assessment while utilizing \emph{rationale} as secondary signals to detect internal contradictions or underdefined specifications. Under this protocol, HLE-Verified is organized into three disjoint subsets: verified gold items, revised items corrected under preserved evaluation objectives, and uncertain items whose validity cannot be conclusively determined.

We position \textbf{HLE-Verified} as dataset infrastructure: not a new benchmark task, but a methodological reliability enhancement that enables more rigorous, interpretable, and reproducible model comparisons. Furthermore, we provide an empirical evaluation of how verification and revision affect downstream model measurements, and we release structured metadata that supports continued community-driven refinement.

\paragraph{Contributions.}
\begin{itemize}
    \item We release \textbf{HLE-Verified}, a verified and revised version of HLE constructed via a transparent component-wise verification protocol and fine-grained error taxonomy, comprising \textbf{668} verified items, \textbf{1,143} revised-and-verified items, and a \textbf{689}-item documented \emph{uncertain} set for structured community refinement.
    \item We introduce a systematic two-stage verification-and-revision framework for post-release benchmark auditing.
    \item We benchmark eight state-of-the-art LLMs on HLE vs.\ HLE-Verified, demonstrating that verification materially alters measured performance (an average \textbf{+7--10} accuracy points overall and \textbf{+30--40} points on items with erroneous problems/answers), and we analyze the relationship between model confidence and problem/answer errors.
\end{itemize}

\section{Background}
\subsection{Why benchmark errors matter}

Benchmarks serve as the primary measurement interface between model development and scientific claims about capability. Let a benchmark contain a fraction $\rho$ of flawed items. Even for small $\rho$, aggregate metrics such as accuracy or pass@k can be meaningfully distorted, particularly when flawed items are not randomly distributed. If such items are concentrated in specific domains, reasoning patterns, or difficulty strata, they may introduce systematic measurement bias \citep{liang2022holistic, gema2025we}, favoring certain model families while penalizing others due to differences in training exposure, reasoning style, or calibration behavior.

Beyond aggregate accuracy, benchmark flaws pose a deeper challenge to interpretability. Many analyses assume a well-defined correspondence between correctness and model confidence, particularly in calibration- or uncertainty-aware evaluations \citep{guo2017calibration, kadavath2022language}. When items are ill-posed, contain incorrect answer keys, or admit multiple valid interpretations, this correspondence becomes ill-defined: models may be penalized for producing valid but unanticipated answers, or rewarded for matching erroneous supervision. In such cases, apparent calibration error or ranking instability may reflect properties of benchmark noise rather than intrinsic model behavior \citep{chao2024make}.

In practice, benchmark flaws commonly manifest in several recurring forms: (1) underspecified or ambiguous problem statements \cite{min2020ambigqa}; (2) incorrect or mismatched answer keys (e.g., unit or format inconsistencies); (3) internal inconsistencies between the provided rationale and the final answer; and (4) reliance on contested facts, implicit conventions, or unstated assumptions. Notably, these failure modes often arise in different components of an evaluation item—the problem statement, the final answer, and the accompanying rationale—suggesting that item validity is multi-dimensional rather than binary.

\subsection{HLE as an evaluation substrate}

Humanity's Last Exam (HLE) has emerged as a widely used benchmark for evaluating frontier language models on challenging, multi-domain questions. In practice, HLE is often reduced to a single scalar performance metric, typically accuracy, occasionally augmented with rationale- or chain-of-thought-based prompting. Implicit in this usage is the assumption that annotation noise is sufficiently small and uniformly distributed such that aggregate comparisons remain meaningful.

However, prior work has demonstrated that large-scale benchmarks may contain systematic annotation errors or unstable evaluation artifacts that meaningfully affect model rankings \citep{gema2025we, prathifkumar2025does}. To date, such concerns have not been systematically quantified for HLE, nor has their impact on evaluation outcomes been rigorously characterized. Consequently, it remains unclear to what extent observed performance differences on HLE reflect genuine capability gaps versus sensitivity to benchmark defects.

This gap motivates a principled re-examination of HLE as an evaluation substrate. Rather than treating benchmark errors as unstructured noise, our work explicitly characterizes item validity, distinguishes different sources of error and uncertainty, and assesses how these factors affect evaluation results. HLE-Verified operationalizes this perspective by making correctness, revision status, and epistemic uncertainty explicit at the dataset level, thereby enabling more reliable and interpretable model comparisons.


\section{Dataset Verification Process and Methods}
This section describes the construction protocol of \textbf{HLE-Verified}, including (i) the end-to-end item pipeline, (ii) the operational definitions of \emph{verified} (gold), \emph{revised}, and \emph{uncertain} items, and (iii) the structured annotation schema used to record error types, revision actions, and item-level epistemic status. The protocol is designed to make benchmark validity explicit, auditable, and component-wise traceable, while preserving the original evaluation intent of HLE wherever possible.

\begin{wrapfigure}{r}{0.5\linewidth}
  \centering
  \vspace{-0.8em}
  \includegraphics[width=1\linewidth]{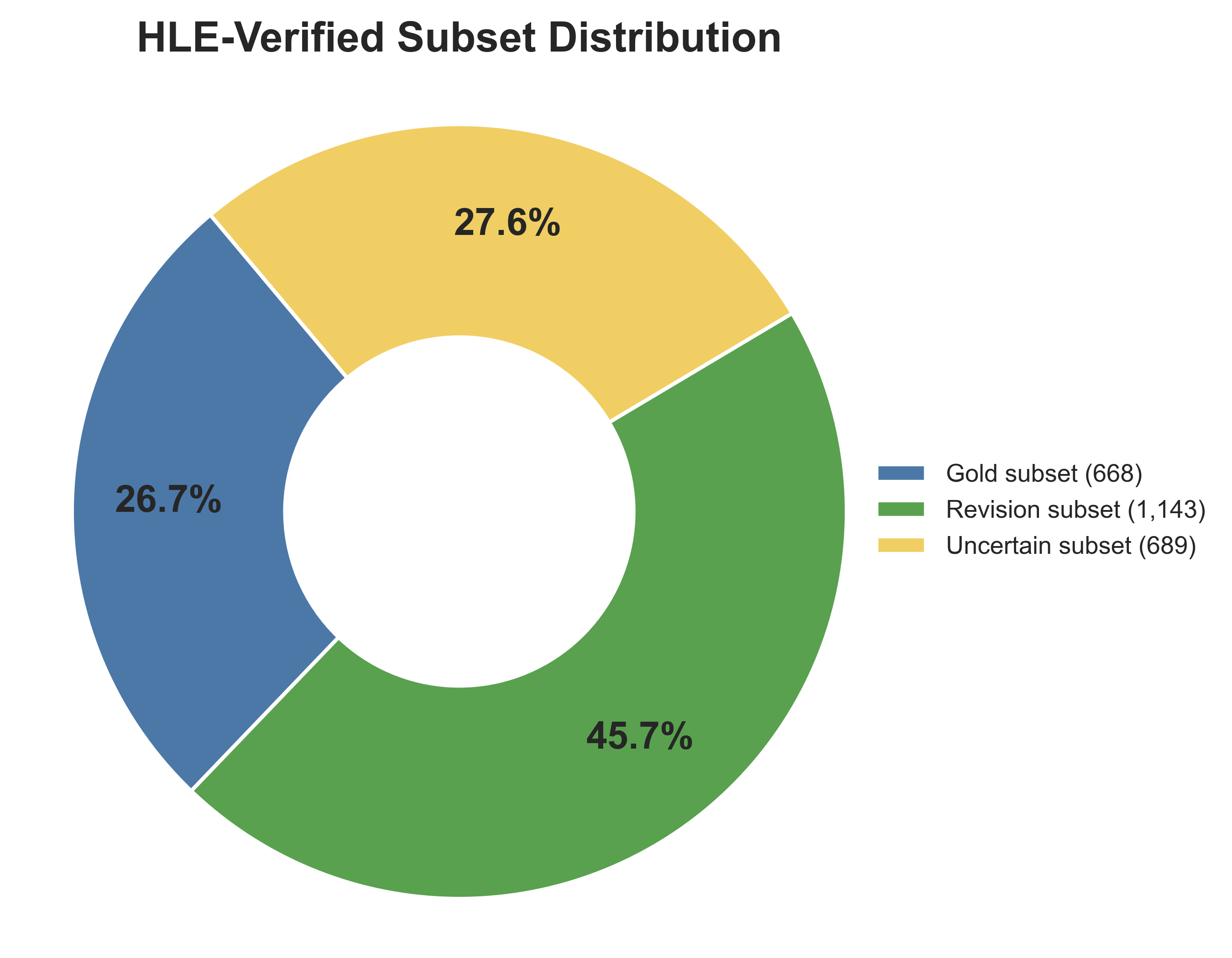}
  \caption{Structural composition of HLE-Verified.}
  \label{fig:hle_verified_composition}
  \vspace{-1em}
\end{wrapfigure}

\subsection{Overview of the pipeline}

Starting from the original HLE collection (2,500 items), we construct a single consolidated benchmark release, \textbf{HLE-Verified}, through a structured two-stage process. \textbf{Stage I} performs component-wise binary verification and produces a high-confidence gold subset. \textbf{Stage II} conducts systematic repair for items judged flawed but repairable, followed by re-verification. Items that remain indeterminate after these procedures are retained as an explicitly documented uncertain subset rather than discarded.

Each item is decomposed into three annotatable components:
(i) the \textbf{problem} (statement plus image, if present),
(ii) the \textbf{final answer}, and
(iii) the \textbf{rationale} (reference solution, when available).

The \textbf{problem} and \textbf{final answer} are primary objects of correctness assessment, defining the evaluand and grading target, while the \textbf{rationale} serves as diagnostic support for detecting inconsistencies, missing assumptions, or explanation defects.

The resulting release comprises three disjoint subsets (Fig~\ref{fig:hle_verified_composition}):
\begin{enumerate}
  \item \textbf{Gold subset} (668 items): validated without modification.
  \item \textbf{Revision subset} (1,143 items): corrected and re-verified under preserved evaluation objectives.
  \item \textbf{Uncertain subset} (689 items): items whose validity remains indeterminate under available expertise and evidence.
\end{enumerate}

Across both stages, we record structured item-level metadata including component-wise validity labels, error-type annotations, revision traces, adjudication notes, and uncertainty descriptors. These fields enable transparent accounting and downstream analyses stratified by defect type and epistemic status.

\subsection{Stage I: component-wise verification}

\begin{figure}[t]
  \centering
  \includegraphics[width=0.98\linewidth]{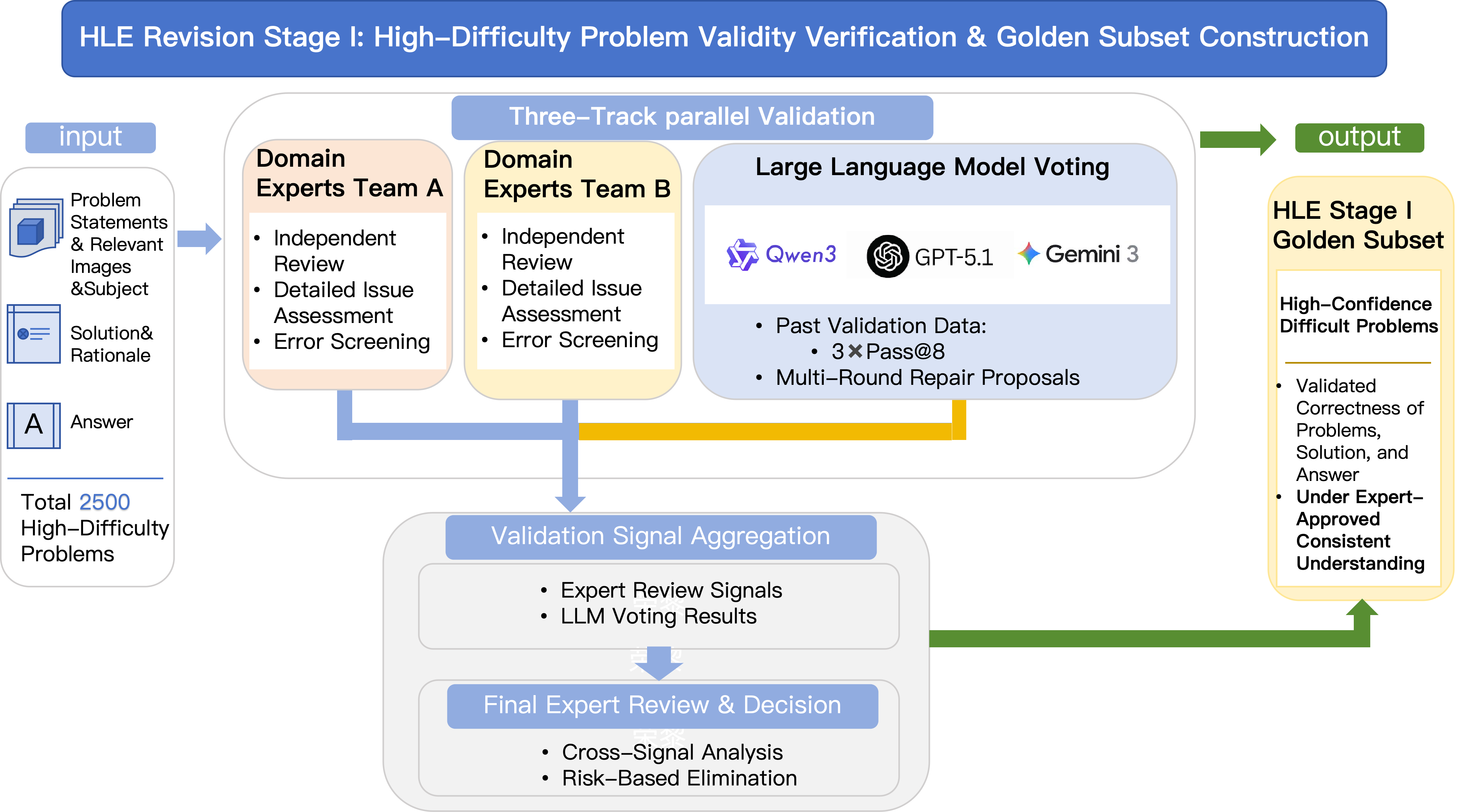}
  \caption{\textbf{HLE Revision Stage I.} High-Difficulty Problem Validity Verification \& Golden Subset Construction}
  \label{fig:stage_I}
\end{figure}
\paragraph{Goal.} Stage I aims to determine whether an item is usable \emph{as originally released}, without performing substantive content edits. The outcome of Stage I is a component-wise binary validity record and a high-confidence gold subset.

\paragraph{Operational definition of validity.}
For each item component, validity is defined as follows:
\begin{itemize}
  \item \textbf{Problem validity:} the statement (and image, if present) is well-posed, self-consistent, and sufficiently specified for a unique or properly qualified solution; assumptions and conventions needed for solving are either explicit or standard within the domain.
  \item \textbf{Answer validity:} the provided final answer is correct with respect to the problem specification, including units, format, and acceptable equivalence classes.
  \item \textbf{Rationale validity:} the reference solution is mathematically/logically sound, internally consistent, and compatible with the final answer (when the rationale is intended as an authoritative reference).
\end{itemize}

\paragraph{Multi-source verification and adjudication.}

Stage I integrates three sources of evidence in a serial–hybrid workflow:

\begin{enumerate}
  \item \textbf{External domain-expert screening.}
  Independent subject-matter reviewers assess problem, answer, and rationale, providing component-wise binary judgments and concise notes.
  \item \textbf{Model-assisted replication checks (pass@8).}
  We invoke multiple frontier multimodal solvers under pass@8 sampling to generate independent solution attempts. Extracted final answers are normalized and compared against the reference answer under a fixed equivalence protocol (numeric tolerance, format normalization, semantic equivalence where applicable). This step provides reproducibility evidence and highlights items exhibiting extreme human–model disagreement.
  \item \textbf{Internal expert adjudication.}
  Internal reviewers synthesize supplier judgments and model-assisted evidence to make conservative inclusion decisions. Items enter the gold subset only when both problem and answer are judged unproblematic and no high-risk ambiguity is identified.
\end{enumerate}

\paragraph{Stage I outcome.}
Stage I yields the \textbf{gold (verified) subset of 668 items} that are retained without modification. All remaining items are carried forward either to Stage II for repair (if flawed but repairable) or to the uncertain pool (if indeterminate or high-risk).

\subsection{Stage II: systematic revision and re-verification}
\begin{figure}[t]
  \centering
  \includegraphics[width=0.98\linewidth]{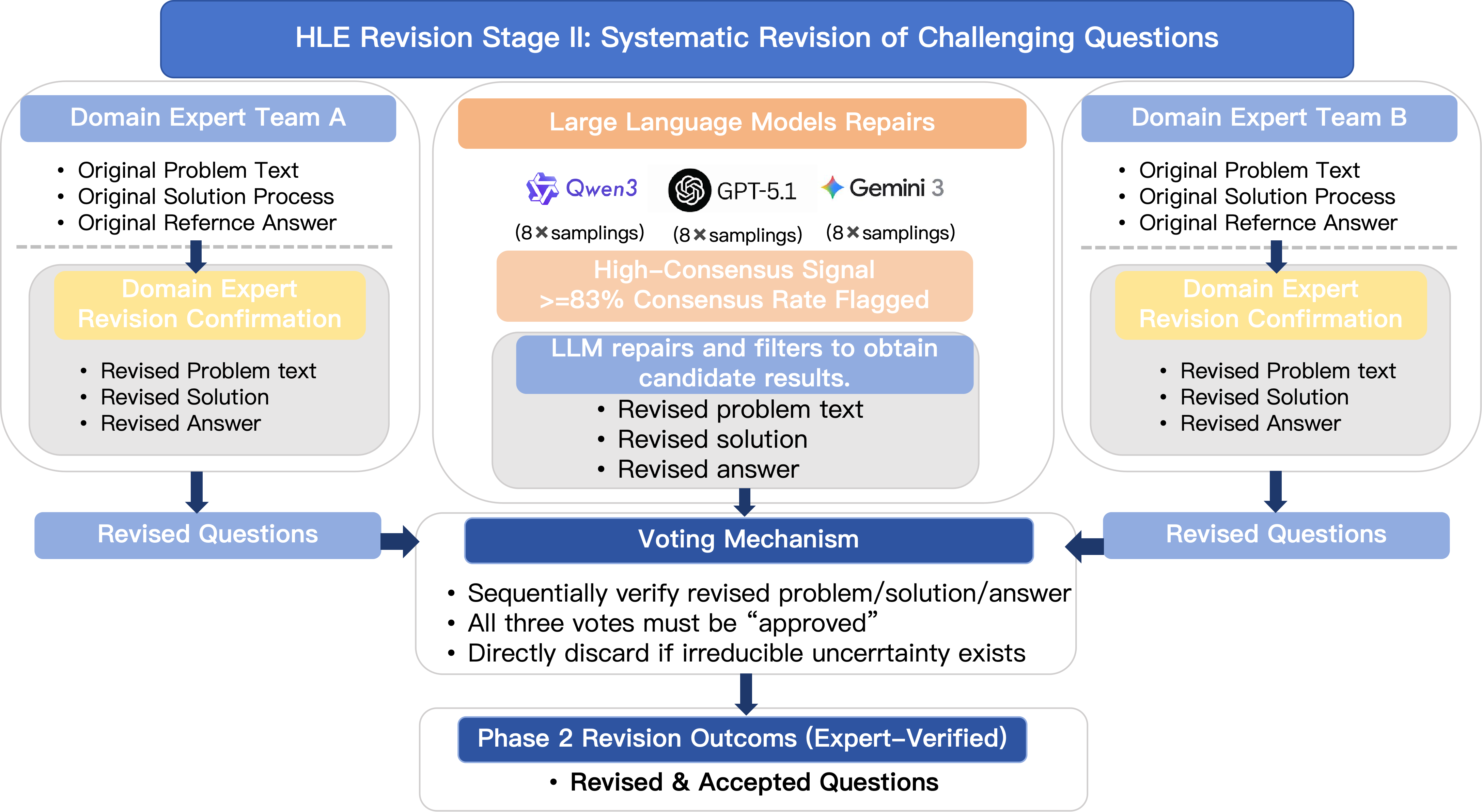}
  \caption{\textbf{HLE Revision Stage II.} Systematic Revision of Challenging Questions}
  \label{fig:stage_II}
\end{figure}

\paragraph{Goal.} Stage II targets items that are judged flawed but \emph{repairable} under the constraint that the original evaluation objective and reasoning target must be preserved. In other words, revisions are corrective rather than creative: they aim to restore well-posedness and correctness without altering what the item is intended to test.

\paragraph{Scope and boundary.} To ensure that revision results are objectively reviewable, Stage II focuses on domains where correctness can be reliably adjudicated and independently reproduced (mathematics, physics, chemistry, biomedicine, and computer science in our current release). Items in domains with weak verification boundaries or subjective conventions are handled conservatively (routed to the uncertain subset).

\paragraph{Revision workflow.} For each candidate item, Stage II generates independent repair proposals followed by expert convergence:
\begin{enumerate}
  \item \textbf{Independent supplier repairs (two-track).}
  Two independent expert teams propose corrective edits following a standardized sequence:
  \emph{Problem Fix} $\rightarrow$ \emph{Solution Fix} $\rightarrow$ \emph{Answer Fix},
  each accompanied by change notes. Items requiring substantive alteration of evaluation intent are marked non-repairable.
  \item \textbf{Model-assisted auxiliary proposals.}
  Multi-model sampling may generate additional repair candidates and stability checks. Model outputs serve as auxiliary evidence and do not replace expert adjudication.
  \item \textbf{Final expert adjudication.}
  Internal experts select or synthesize a canonical repaired version under the principles of objective preservation, correctness, and minimal necessary edits. Items remaining ambiguous or unverifiable are routed to the uncertain subset.
\end{enumerate}

\paragraph{Stage II outcome.}
Stage II yields the \textbf{revision subset of 1,143 items} that are corrected and re-verified as suitable for evaluation. Each revised item includes structured revision metadata, including which components were changed, what error types were addressed, and brief adjudication notes sufficient for auditing.

\subsection{Uncertain subset: epistemic status and documentation}

After verification and revision, 689 items remain uncertain. Rather than discarding them, we retain these items as an explicit epistemic category, reflecting cases where validity cannot be established with sufficient confidence. An item is assigned to the uncertain subset when resolution would require non-standard assumptions, authoritative external references, unresolved expert disagreement, or domain knowledge beyond the current verification scope.

Each uncertain item includes structured documentation:
\begin{itemize}
  \item an \textbf{uncertainty source label},
  \item and a \textbf{required expertise tag} indicating the type of specialist input needed for resolution.
\end{itemize}
This design separates capability limits from benchmark indeterminacy and enables principled community refinement.

\subsection{Component-wise annotation framework and defect taxonomy}
\label{sec:annotation_schema}

All HLE-Verified items are annotated at the component level to enable localized defect attribution, structured revision tracking, and reproducible auditing. Each item is decomposed into three semantically distinct components: the \emph{problem statement} (\textbf{Problem}), the \emph{reference rationale/solution} (\textbf{Rationale}), and the \emph{final answer} (\textbf{Answer}). Defect labels and revision records are assigned to the specific component in which the issue originates, rather than treating item validity as a monolithic binary property.

Each defect category corresponds to a violation of component-level validity constraints, reflecting that item validity is structured and multi-dimensional. We define a taxonomy comprising \textbf{19 categories} in total, organized as \textbf{5} problem-level errors, \textbf{10} rationale-level errors, and \textbf{4} answer-level errors. The taxonomy serves as the unified basis for (i) statistical reporting of defect prevalence, (ii) analysis of repair patterns in the revised subset, and (iii) public release of structured item-level metadata.

\begin{figure}[t]
  \centering
  \includegraphics[width=0.98\linewidth]{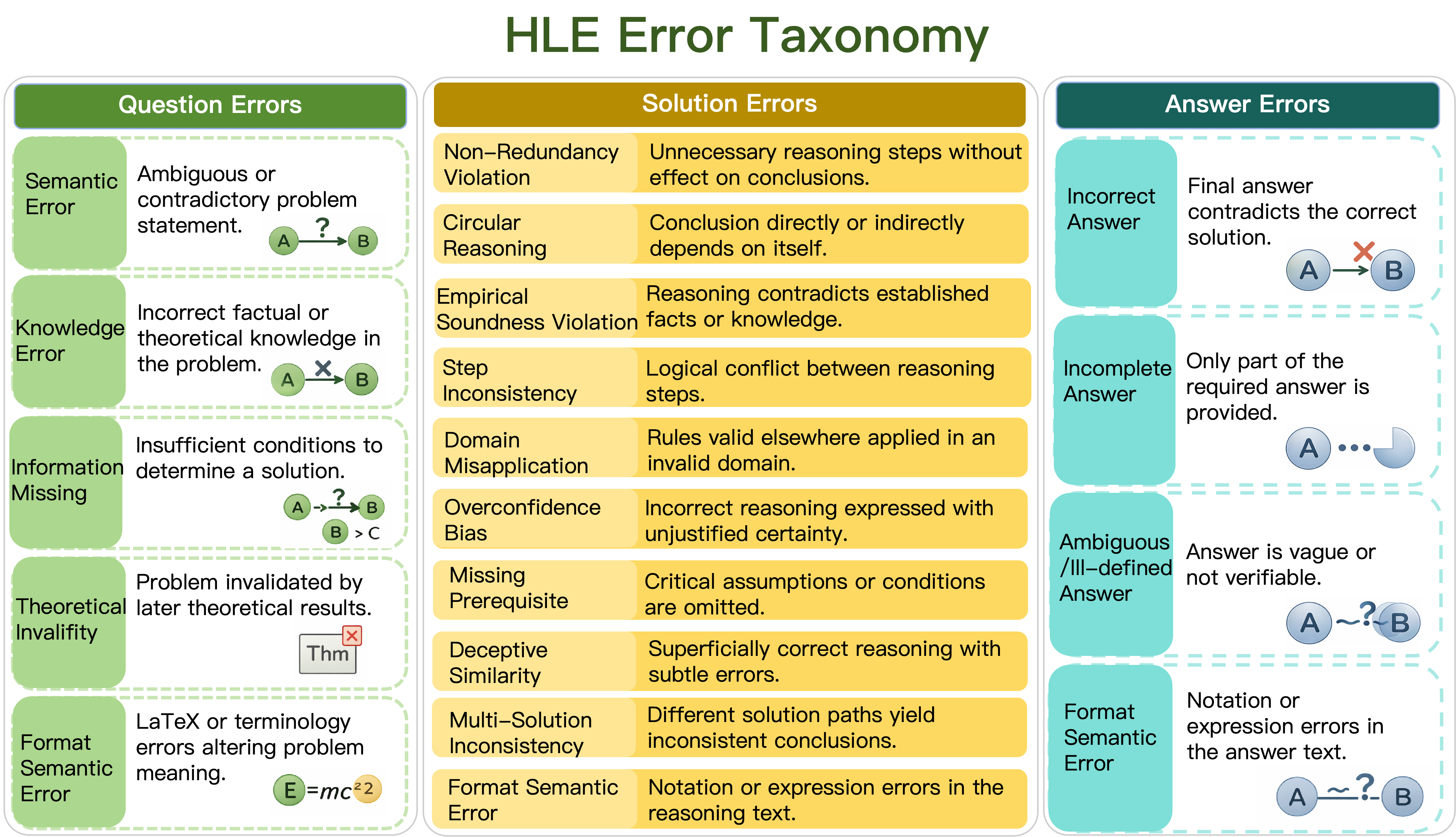}
  \caption{HLE Component-wise Defect Taxonomy}
  \label{fig:error_types}
\end{figure}

\paragraph{Problem-level defects.}
Problem-level defects arise from flaws in task specification that compromise interpretability, solvability, or semantic faithfulness.

\begin{itemize}
  \item \textbf{(Q1) Semantic Error.} The statement is ambiguous, contradictory, or underspecified, admitting multiple interpretations.
  \item \textbf{(Q2) Knowledge Error.} The statement contains incorrect factual premises or misuse of established domain knowledge.
  \item \textbf{(Q3) Missing Information.} Essential constraints or assumptions required for solvability or uniqueness are omitted.
  \item \textbf{(Q4) Theoretical Invalidity.} The statement is invalid under accepted theory, supported by explicit corrective evidence.
  \item \textbf{(Q5) Format Semantic Error (Problem).} Notation, LaTeX, or terminology defects that distort or obscure intended meaning.
\end{itemize}

\paragraph{Rationale-level defects.}
Rationale-level defects characterize failures in the \emph{reference reasoning chain} provided by the benchmark. These categories apply to the authoritative solution text rather than to model-generated reasoning.

\begin{itemize}
  \item \textbf{(S1) Non-Redundancy Violation.} Redundant reasoning steps that reduce minimality or auditability.
  \item \textbf{(S2) Circular Reasoning.} The conclusion depends on itself directly or indirectly.
  \item \textbf{(S3) Empirical Soundness Violation.} Steps contradict established facts or accepted knowledge.
  \item \textbf{(S4) Step Inconsistency.} Logical conflicts among intermediate steps.
  \item \textbf{(S5) Domain Misapplication.} Misuse of rules or theorems outside their valid scope.
  \item \textbf{(S6) Overconfidence Bias.} Incorrect content stated with unjustified certainty.
  \item \textbf{(S7) Missing Prerequisite.} Critical assumptions or conditions omitted while proceeding with derivation.
  \item \textbf{(S8) Deceptive Similarity.} Superficially plausible reasoning containing subtle structural corruption.
  \item \textbf{(S9) Multi-Solution Inconsistency.} Inconsistency across legitimate solution paths or case analyses.
  \item \textbf{(S10) Format Semantic Error (Rationale).} Symbol, LaTeX, unit, or terminology issues impairing verifiability or semantic alignment.
\end{itemize}

\paragraph{Answer-level defects.}
Answer-level defects concern correctness, completeness, and verifiability of the final output.

\begin{itemize}
  \item \textbf{(A1) Incorrect Answer.} The final answer is inconsistent with the correct solution (e.g., sign/value/boolean inversion).
  \item \textbf{(A2) Incomplete Answer.} Required cases, qualifiers, or multi-part conclusions are missing.
  \item \textbf{(A3) Ambiguous / Ill-defined Answer.} The answer is non-verifiable due to vagueness or mismatch with required output form.
  \item \textbf{(A4) Format Semantic Error (Answer).} Expression-level defects (e.g., LaTeX errors, incorrect symbols, unit inconsistencies) that impair interpretability without necessarily altering conceptual content.
\end{itemize}

\paragraph{Revision and epistemic annotations.}
For items in the revised subset, we record component-level modification indicators 
(\texttt{problem\_fix}, \texttt{solution\_fix}, \texttt{answer\_fix}) together with concise change notes. Revisions are performed under the constraint that the \emph{original evaluation objective and reasoning target} are preserved; corrections restore validity and internal consistency without redefining the underlying capability being assessed.

Each item additionally includes an epistemic status label (\emph{verified}, \emph{revised}, or \emph{uncertain}) and structured uncertainty descriptors where applicable. These fields make correction history, defect attribution, and indeterminacy explicit at the dataset level, enabling transparent auditing and stratified evaluation analyses.


\begin{figure}[t]
  \centering
  \includegraphics[width=0.7\linewidth]{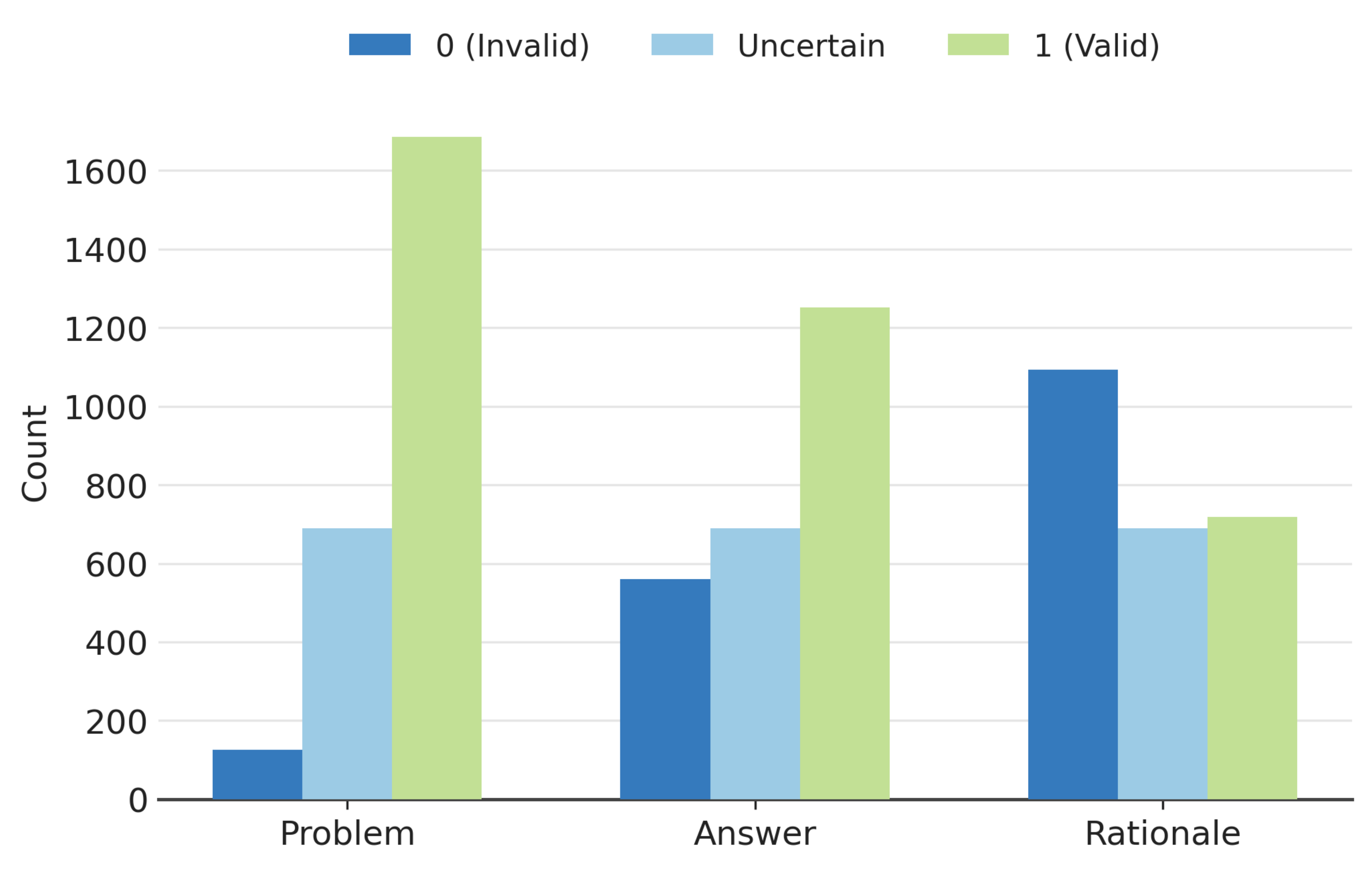}
  \caption{\textbf{Overall annotation outcomes on the problematic data of HLE.}
We report the counts of three labels---valid (1), invalid (0), and uncertain---for each annotatable component: \emph{Problem}, \emph{Answer}, and \emph{Rationale}. The results show clear component-wise differences, with substantially more invalid/uncertain cases in \emph{Answer} and especially \emph{Rationale} than in \emph{Problem}.}
  \label{fig:validity_distribution}
\end{figure}

\begin{figure}[t]
  \centering
  \includegraphics[width=0.98\linewidth]{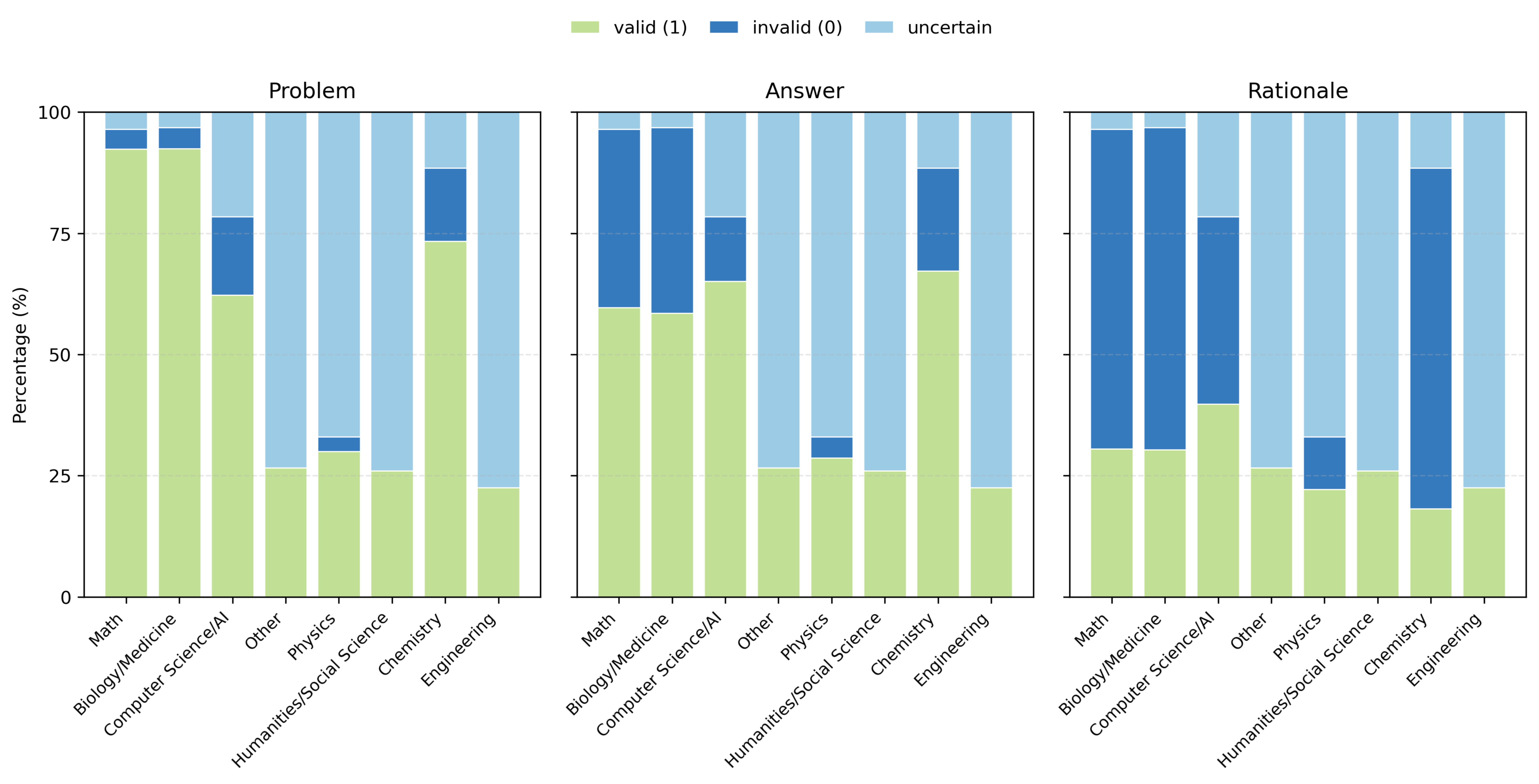}
  \caption{\textbf{Annotation outcome distribution across subject categories.}
For each subject category, we visualize the percentage of valid (1), invalid (0), and uncertain labels for the \emph{Problem}, \emph{Answer}, and \emph{Rationale} components. The label composition varies by domain, while the \emph{Rationale} component consistently exhibits the lowest validity and the highest uncertainty/invalidity across categories.}
  \label{fig:validity_by_category}
\end{figure}

\section{Dataset Statistical Analysis}

We report a statistical analysis of the \emph{problematic subset} of HLE-Verified, defined as items that did not enter the gold subset in Stage I. Our analysis focuses on (i) component-wise annotation outcomes (Problem, Answer, Rationale), and (ii) cross-domain variation in defect patterns. We further summarize recurring failure modes to contextualize the distributional findings.

\subsection{Component-wise annotation outcomes}

Each HLE instance is decomposed into three annotatable components—\textbf{Problem}, \textbf{Answer}, and \textbf{Rationale}—with labels \textbf{valid} (1), \textbf{invalid} (0), or \textbf{uncertain}. Figure~\ref{fig:validity_distribution} presents the aggregate distribution across the problematic subset.


Across components, the \textbf{Problem} field exhibits the highest reliability. The majority of problem statements are structurally valid, and explicit structural errors constitute only a small fraction of cases. This indicates that most instances are well-posed at the task-description level. In contrast, reliability decreases at the \textbf{Answer} level. Only about half of the answers are fully valid, with a substantial portion either incorrect or difficult to verify. This suggests that correctness mismatches and solution inconsistencies are a major source of dataset noise. The most pronounced degradation appears in the \textbf{Rationale} component. Invalid rationales outnumber valid ones, indicating that reasoning chains frequently contain logical gaps, unsupported inferences, or incorrect derivations—even when the corresponding problem statement appears acceptable.

Importantly, the \textbf{uncertain} label constitutes a non-trivial fraction across all components. These cases typically arise from underspecified assumptions, ambiguous notation, incomplete derivations, or dependence on implicit conventions. The prevalence of uncertainty underscores that problematicness in HLE is not limited to outright incorrectness but often involves epistemic indeterminacy.

\subsection{Cross-domain variation}

Figure~\ref{fig:validity_by_category} further disaggregates component-wise validity by subject category, revealing pronounced cross-domain differences.

For the \textbf{Problem} component, formal scientific domains exhibit markedly higher structural reliability. \textbf{Math} and \textbf{Biology/Medicine} both exceed 92\% validity, while \textbf{Chemistry} remains comparatively strong at 73.3\%. In contrast, \textbf{Physics} shows only 30.0\% valid problem statements, and \textbf{Engineering}, \textbf{Humanities/Social Science}, and \textbf{Other} fall further to the 22–27\% range. Notably, in these lower-validity domains, the majority of remaining cases are labeled \textbf{uncertain} (roughly 67–77\%), rather than explicitly invalid, indicating verification ambiguity rather than structural error.

At the \textbf{Answer} level, domain gaps persist. \textbf{Computer Science/AI} (65.1\%) and \textbf{Chemistry} (67.3\%) achieve the highest answer validity rates. \textbf{Math} (59.6\%) and \textbf{Biology/Medicine} (58.6\%) remain above 50\%, but both exhibit substantial invalid proportions (36.8\% and 38.2\%, respectively), suggesting that answer correctness in these domains is typically decidable and frequently incorrect rather than ambiguous. By contrast, \textbf{Physics}, \textbf{Engineering}, \textbf{Humanities/Social Science}, and \textbf{Other} remain below 30\% valid, with uncertainty again dominating (approximately two-thirds to three-quarters of cases).

The divergence becomes most pronounced for the \textbf{Rationale} component. In \textbf{Math} and \textbf{Biology/Medicine}, invalid rationales dominate (65.9\% and 66.4\%, respectively), and \textbf{Chemistry} reaches an even higher invalid rate of 70.3\%. These domains therefore display explicit logical or derivational errors even when problems are well-formed. In contrast, \textbf{Engineering}, \textbf{Humanities/Social Science}, and \textbf{Other} show extremely high uncertainty rates (approximately 73–77\%) with virtually no explicit invalidation, reflecting verification difficulty rather than clear logical contradiction. \textbf{Physics} occupies an intermediate position, with high uncertainty (67.0\%) but a non-trivial invalid share (10.9\%).

Overall, these findings reinforce two structural observations. First, problematicness in HLE is rarely concentrated in problem statements alone; answer- and rationale-level defects account for the majority of reliability degradation. Second, domains differ not only in defect prevalence but in defect type: some disciplines exhibit predominantly incorrect content, while others exhibit epistemic indeterminacy due to verification complexity. These patterns justify a component-wise and epistemically explicit verification framework, as aggregate item-level labeling would obscure substantial heterogeneity in defect structure.

\subsection{Component-wise Defect Distribution Across Subjects}
\label{sec:defect_distribution_analysis}

We next analyze the distribution of defect categories within the revised subset of HLE-Verified using the 19-category component-wise taxonomy. Rather than reporting raw counts alone, we examine the \emph{relative proportions} of defect types within each component, both globally and across subjects (Figure~\ref{fig:error_dist_subjects}).

\paragraph{Global component-level proportions.}

When aggregating across subjects, three structural patterns emerge.





\textbf{Answer-level defects.}
Across all five domains, \emph{Incorrect Answer} (type 1) is consistently the dominant answer-level defect. Its within-component proportion ranges from 69.4\% (Chemistry) to 97.2\% (Biology/Medicine), with all subjects exceeding 70\%. This indicates that answer-level failures are primarily deterministic correctness violations rather than ambiguity, partial specification, or formatting issues.

\textbf{Rationale-level defects.}
Rationale defects are more heterogeneous across subjects. In Mathematics (40.1\%) and Biology/Medicine (45.0\%), type 3 defects (structural incompleteness) dominate, whereas Chemistry (40.0\%) and Computer Science (63.3\%) are led by type 10 defects (format-induced semantic errors). Physics shows a flatter distribution, with type 3 accounting for 25.0\% of rationale defects. Overall, rationale instability arises primarily from missing intermediate structure and representation misalignment rather than classical logical contradictions.

\textbf{Problem-level defects.}
Problem-level defect patterns are domain-sensitive. Mathematics (40.0\%), Chemistry (56.5\%), and Computer Science (94.6\%) are dominated by type 5 defects (format semantic errors), whereas Physics is led by type 1 (57.1\%) and Biology/Medicine by type 2 (33.3\%). The near absence of dominant theoretical-invalidity categories suggests that most flawed items originate from specification or representation-level imprecision rather than fundamentally incorrect task concepts.

Taken together, these proportions indicate that reliability degradation in HLE is primarily driven by answer-key instability and representation-level specification errors, with pronounced disciplinary heterogeneity in defect structure.

\paragraph{Structural implications.}

Across domains, several robust regularities emerge:


\begin{enumerate}
    \item \textbf{Incorrect final answers dominate answer-level instability.}
    Deterministic answer-key errors constitute the principal failure mode across all subjects.
    \item \textbf{Rationale instability is primarily structural or representational.}
    Missing intermediate steps and format-level semantic misalignment outweigh classical logical contradictions.
    \item \textbf{Representation sensitivity is discipline-dependent.}
    Format semantic errors play a particularly large role in Chemistry and Computer Science.
    \item \textbf{Conceptual invalidity is not the dominant driver.}
    Most defects arise from specification or answer-key instability rather than from fundamentally incorrect task concepts.
\end{enumerate}

These findings suggest that defect patterns in HLE exhibit systematic structure rather than uniform random noise. Reliability risks are component-specific and discipline-sensitive, with answer correctness and rationale completeness constituting the dominant bottlenecks. This empirical structure justifies a component-wise verification protocol and cautions against treating benchmark noise as uniform or negligible.

\begin{table*}[t]
\centering
\small
\caption{Dominant defect type (Top-1) by subject and component. 
Percentages denote within-component ratios.}
\label{tab:top1_defects}
\begin{tabular}{lccc}
\toprule
\textbf{Subject} 
& \textbf{Problem (Top-1)} 
& \textbf{Rationale (Top-1)} 
& \textbf{Answer (Top-1)} \\
\midrule

Mathematics
& Format Semantic Error (40.0\%) 
& Information Missing (40.1\%) 
& Incorrect Answer (89.4\%) \\

Physics 
& Semantic Error (57.1\%) 
& Information Missing (25.0\%) 
& Incorrect Answer (80.0\%) \\

Chemistry 
& Format Semantic Error (56.5\%) 
& Format Semantic Error (40.0\%) 
& Incorrect Answer (71.4\%) \\

Biology/Medicine 
& Knowledge Error (33.3\%) 
& Information Missing (45.0\%) 
& Incorrect Answer (97.2\%) \\

Computer Science 
& Format Semantic Error (94.6\%) 
& Format Semantic Error (63.3\%) 
& Incorrect Answer (78.1\%) \\

\bottomrule
\end{tabular}
\end{table*}

Table~\ref{tab:top1_defects} summarizes the dominant defect type (Top-1) for each subject and component. The table highlights a consistent asymmetry: answer-level defects are overwhelmingly dominated by \emph{Incorrect Answer} across all subjects, whereas problem- and rationale-level dominant defects vary by discipline. In particular, format-induced semantic errors dominate Computer Science, while structural incompleteness is most prevalent in Mathematics and Physics. These patterns reinforce the component-specific and domain-sensitive nature of reliability degradation in HLE.


\subsection{Case Studies}
\label{sec:case_studies}

To complement the quantitative findings, we present representative case studies illustrating structural defect patterns identified in HLE-Verified. These examples demonstrate that benchmark failures are not isolated anomalies but manifestations of systematic component-level weaknesses. We focus on three recurring families aligned with our component-wise analysis:

(1) \textbf{Answer-level errors} (incorrect conclusions, unit/format mismatches, inconsistency with the problem specification),  
(2) \textbf{Rationale-level defects} (missing prerequisites, circular reasoning, empirical violations, internally inconsistent derivations), and  
(3) \textbf{Uncertainty-inducing ambiguity} (underspecified assumptions or multiple valid interpretations).

These illustrate why component-wise auditing is essential: an item may present a well-formed problem statement yet remain unsuitable for evaluation due to an incorrect answer key or a structurally non-verifiable rationale. Below we present representative canonical cases selected by domain experts; the complete revision artifacts and detailed correction records are available in the open-sourced HLE-Verified benchmark.

\bigskip

\begin{tcolorbox}[title=Case 1: Theoretical–Implementation Confusion (Computer Science)]
\textbf{Defect Tags:} S3 Empirical Soundness Violation + A1 Incorrect Answer

\textbf{Error Mechanism.}  
A speculative decoding sanity-check asked for the expected acceptance rate when the same model serves as both draft and target.  
The original solution claimed the rate should be \emph{less than 1} due to GPU kernel differences (GEMV vs.\ GEMM) producing slightly different logits.  
This conflated implementation-level numerical variability with a theoretical property of the algorithm.

\textbf{Revision.}  
Separating theory from hardware artifacts yields:
\[
r=\min\!\Bigl(1,\frac{P_{\text{target}}(t)}{P_{\text{draft}}(t)}\Bigr)=1.
\]
The correct answer is \emph{exactly 1} under identical models.

\textbf{Significance.}  
Prevents elevating engineering noise into theoretical reasoning—critical for ML benchmark integrity.
\end{tcolorbox}





\begin{tcolorbox}[title=Case 2: Numerical Inconsistency in Stoichiometric Constraint (Chemistry)]
\textbf{Defect Tags:} Q2 Knowledge Error + S3 Empirical Soundness Violation

\textbf{Error Mechanism.}  
Incorrect molecular mass constants rendered the trimer-plus-lipid relation mathematically inconsistent, making the system unsatisfiable.

\textbf{Revision.}  
Corrected constants restore internal coherence:
\[
3 \times \text{protein} + 3 \times \text{lipid} = 101553\,\text{Da}.
\]
The lipid mass (1501 Da) aligns with cardiolipin.

\textbf{Significance.}  
Transforms an incoherent specification into a structurally valid biochemical reasoning task.
\end{tcolorbox}

\begin{tcolorbox}[title=Case 3: Sign Inconsistency in Exciton Energy (Physics)]
\textbf{Defect Tags:} S3 Empirical Soundness Violation + A1 Incorrect Answer

\textbf{Error Mechanism.}  
Confusion among bandgap energy, resonance peak, and binding energy introduced a negative “Rydberg energy,” violating physical interpretation.

\textbf{Revision.}
\[
E_b(1)=2~\text{eV}, \quad Ry^*=0.5~\text{eV},
\]
\[
E_b(3)=0.08~\text{eV}.
\]

\textbf{Significance.}  
Restores physical coherence and eliminates brittle sign-convention errors.
\end{tcolorbox}









\bigskip

Across domains, failures cluster in recurring structural patterns: theoretical–implementation confusion, constraint omission, numerical incoherence, sign inconsistency, invariant violation, and conceptual mislocalization. These are structural validity failures rather than cosmetic defects. Stage~II revisions therefore restore logical soundness, theoretical alignment, and domain consistency, reinforcing that benchmark reliability depends on component-level integrity rather than superficial correctness.


\section{Experimental Results}
\subsection{Experimental Setup}
\paragraph{Models.}
We evaluate the following frontier models: \textbf{GPT-5.2-Thinking}~\citep{openai2025_introducing_gpt_5_2}, \textbf{Gemini3-Pro-Preview}~\citep{googledeepmind_gemini_pro}, \textbf{Claude-Opus4.5}~\citep{anthropic_claude_opus_4_5}, \textbf{Claude-Opus4.6}~\citep{anthropic_claude_opus_4_6}, \textbf{Grok-4.1 (fast-reasoning)}~\citep{xai2025_grok_4_1_fast}, \textbf{DeepSeek-V3.2-Thinking}~\citep{liu2025deepseek}, \textbf{Qwen3-Max-Thinking}~\citep{qwen3maxthinking}, \textbf{Qwen3.5-Plus}~\citep{qwen35blog}. All models were evaluated using the same system prompt recommended by the HLE official guidelines, along with each model’s own default recommended decoding configuration. To reduce variance from stochastic decoding, we run five independent rollouts per item and report avg5 accuracy, i.e., the average correctness across the five sampled completions (without additional post-hoc selection unless otherwise stated).

\paragraph{Metrics.}
We report (i) \textbf{Accuracy (Acc)} and (ii) \textbf{Calibration Error (Cali Err)}. Calibration error is computed from the model's self-reported confidence and the binary correctness label: we parse each response-level confidence into $c\in[0,1]$ and compute
\[
\text{Cali Err} = 100\times \mathrm{calib\_err}(c,y;\,p=2,\beta=100),
\]
where $y\in\{0,1\}$ indicates whether the item is answered correctly, and $\mathrm{calib\_err}$ is a smoothed $L_2$ miscalibration estimator with smoothing parameter $\beta$. Note that this $\mathrm{calib\_err}$ estimator follows the official HLE evaluation code.

\paragraph{Evaluation Datasets.}
We report results on (i) the \textbf{Full Set} (Raw HLE Full vs.\ Revised HLE-Verified Full) to quantify the end-to-end impact on the benchmark, and (ii) a \textbf{Revised Subset} comparison (Raw HLE Subset vs.\ Revised HLE-Verified Subset) limited to items that were edited/flagged during our verification process, which isolates the effect of flawed items and their corrections. Note that, to ensure comprehensive revisions, we also corrected the original rationales and released them to the open-source community. However, under the standard HLE benchmark evaluation, items with errors only in the rationale do not affect the final scores, since mainstream evaluation uses only the problem and answer. Therefore, the \textbf{Revised Subset} in this experimental section includes only items for which at least one of the problem/answer fields was revised. Unless otherwise specified, all evaluation sets used in this section are the text-only subset of HLE.


\subsection{Main Evaluation: HLE vs.\ HLE-Verified}
Table~\ref{tab:hle_results} summarizes performance on the original benchmark and on HLE-Verified. We focus on the accuracy shift
\[
\Delta \text{Acc} = \mathrm{Acc}(\text{HLE-Verified}) - \mathrm{Acc}(\text{HLE}),
\]
computed either on the Full Set or on the Revised Subset.

\paragraph{Large accuracy gains on revised items.}
On the \textbf{Revised Subset}, all models exhibit substantial positive shifts, indicating that flawed items in the original benchmark systematically suppress measured performance. Specifically, accuracy increases by: Gemini-3-pro (+29.94), GPT-5.2 (+38.04), Claude-Opus4.5 (+32.94), Grok-4.1 fast-reasoning (+34.82), Claude-Opus4.6 (+30.13), and DeepSeek-V3.2 (+39.58) percentage points. The magnitude of these shifts implies that a non-trivial fraction of ``errors'' on raw HLE are attributable to benchmark issues rather than model capability.

\paragraph{Calibration improves after verification.}
Calibration error decreases consistently after revision on the same subset (e.g., GPT-5.2: 63$\rightarrow$28; DeepSeek-V3.2: 70$\rightarrow$28; Grok-4.1: 83$\rightarrow$47), suggesting that flawed items can also distort confidence-based evaluation by inducing confidently incorrect (or otherwise miscalibrated) responses. In contrast, HLE-Verified yields a more faithful estimate of the confidence--correctness relationship.

\paragraph{Full-set results.}
On the \textbf{Full Set}, we also observe consistent gains after verification, though smaller in magnitude than on the Revised Subset, as expected since most items are unchanged. Specifically, accuracy increases by: Gemini-3-pro (+7.58), GPT-5.2 (+9.95), Claude-Opus4.5 (+8.68), Grok-4.1 fast-reasoning (+9.26), Claude-Opus4.6 (+7.75), DeepSeek-V3.2 (+10.79), and Qwen3-Max-Thinking (+8.92) percentage points. Calibration error likewise decreases across models (e.g., GPT-5.2: 45$\rightarrow$36; DeepSeek-V3.2: 56$\rightarrow$46; Grok-4.1: 73$\rightarrow$63), indicating that HLE-Verified not only raises measured accuracy but also yields more reliable confidence estimates at benchmark scale.
\begin{table*}[t]
\centering
\caption{Results on Full Set and Revised Subset.}
\resizebox{\textwidth}{!}{%
\begin{tabular}{lcccccccccc}
\toprule
& \multicolumn{5}{c}{Full Set} & \multicolumn{5}{c}{Revised Subset} \\
\cmidrule(lr){2-6}\cmidrule(lr){7-11}
Dataset / Model-Metrics
& \multicolumn{2}{c}{Raw: HLE Full}
& \multicolumn{2}{c}{Revised: HLE-Verified Full}
& \multicolumn{1}{c}{$\Delta$}
& \multicolumn{2}{c}{Raw: HLE Subset}
& \multicolumn{2}{c}{Revised: HLE-Verified Subset}
& \multicolumn{1}{c}{$\Delta$} \\
\cmidrule(lr){2-3}\cmidrule(lr){4-5}\cmidrule(lr){6-6}
\cmidrule(lr){7-8}\cmidrule(lr){9-10}\cmidrule(lr){11-11}
& Acc & Cali Err & Acc & Cali Err & $\Delta$
& Acc & Cali Err & Acc & Cali Err & $\Delta$ \\
\midrule
Gemini3-pro         & 40.42 & 56 & 48.2  & 49 &  7.78 & 18.99 & 74 & 48.93 & 45 & 29.94 \\
GPT-5.2-High       & 33.35 & 45 & 43.3  & 36 &  9.95 & 14.44 & 63 & 52.48 & 28 & 38.04 \\
Claude-Opus4.5     & 30.00 & 55 & 38.8  & 46 &  8.80  & 14.20 & 70 & 48.16 & 39 & 33.96 \\
Grok 4.1-fast-reasoning
                   & 19.94 & 73 & 29.0  & 63 &  9.06 &  8.25 & 83 & 43.07 & 47 & 34.82 \\
Claude-Opus4.6     & 38.95 & 40 & 46.8  & 32 &  7.85 & 20.03 & 59 & 50.16 & 27 & 30.13 \\
Deepseek-V3.2      & 24.90  & 56 & 36.4  & 46 & 11.50  &  7.87 & 70 & 47.45 & 28 & 39.58 \\
Qwen3-Max-Thinking & 30.00  & 66 & 38.2  & 57 &  8.20  & 14.2  & 79 & 48.48 & 44 & 34.28 \\
\bottomrule
\end{tabular}%
}
\label{tab:hle_results}
\end{table*}

\begin{figure}[t]
  \centering
  \includegraphics[width=0.48\linewidth]{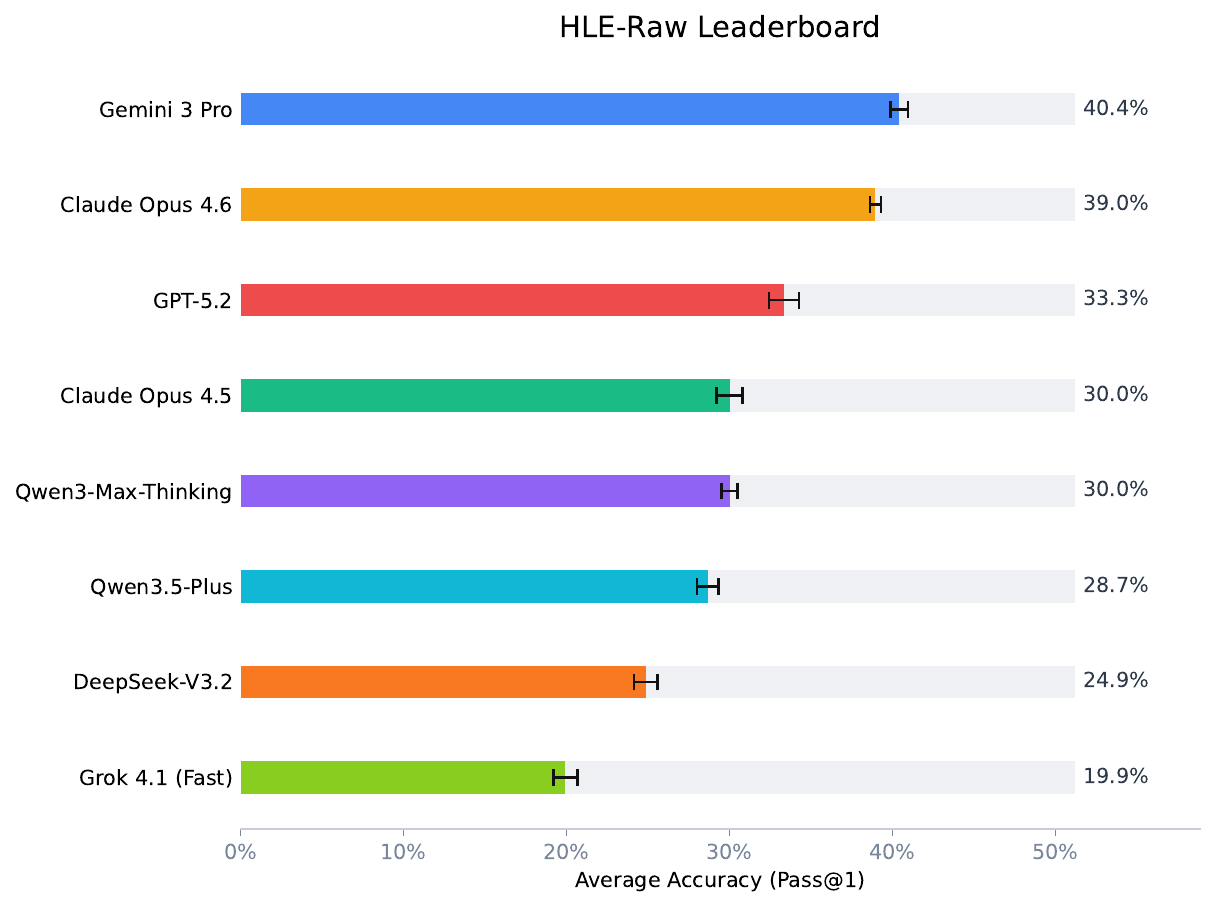}
  \includegraphics[width=0.48\linewidth]{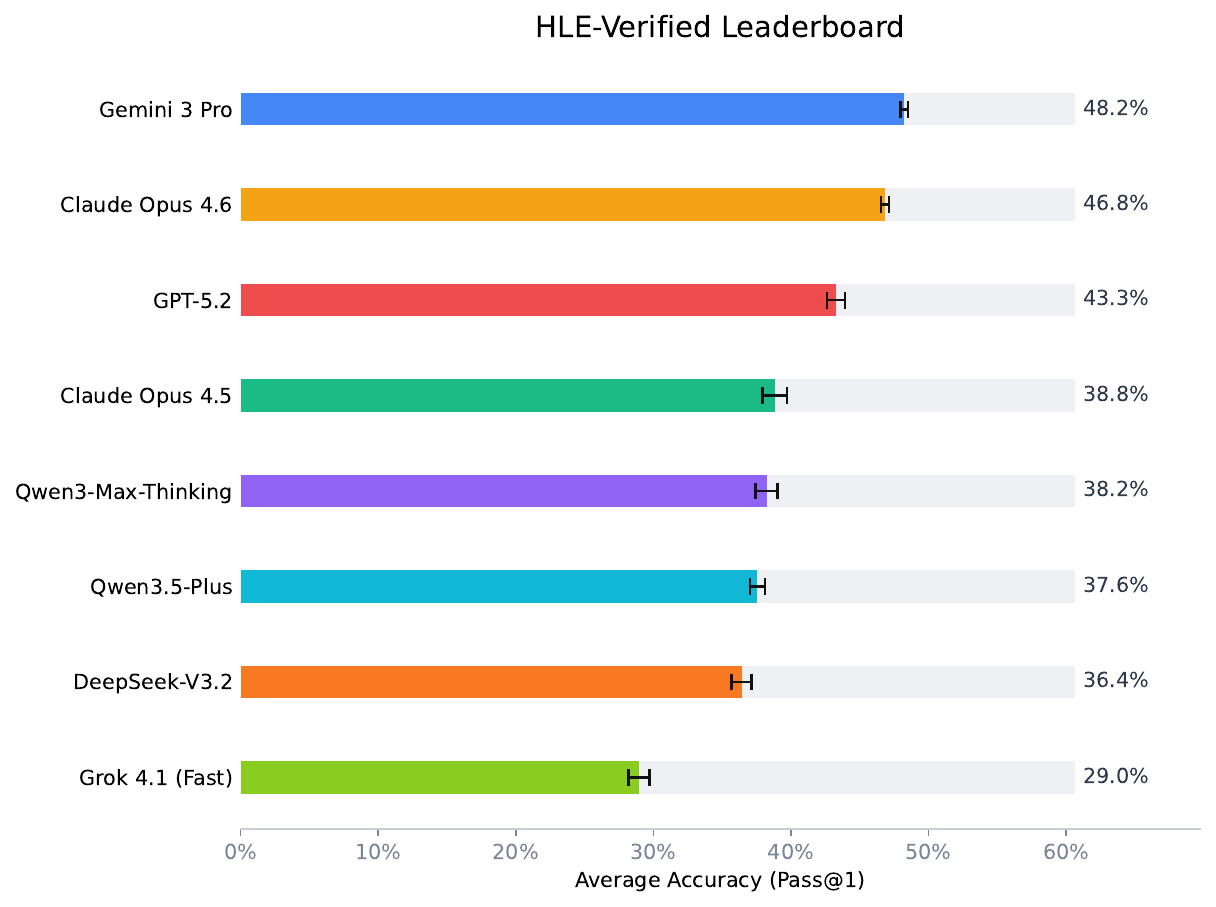}
  \caption{HLE Raw LeaderBoard vs HLE-Verified LeaderBoard}
  \label{fig:leaderboard_full}
\end{figure}

\begin{figure}[t]
  \centering
  \includegraphics[width=0.7\linewidth]{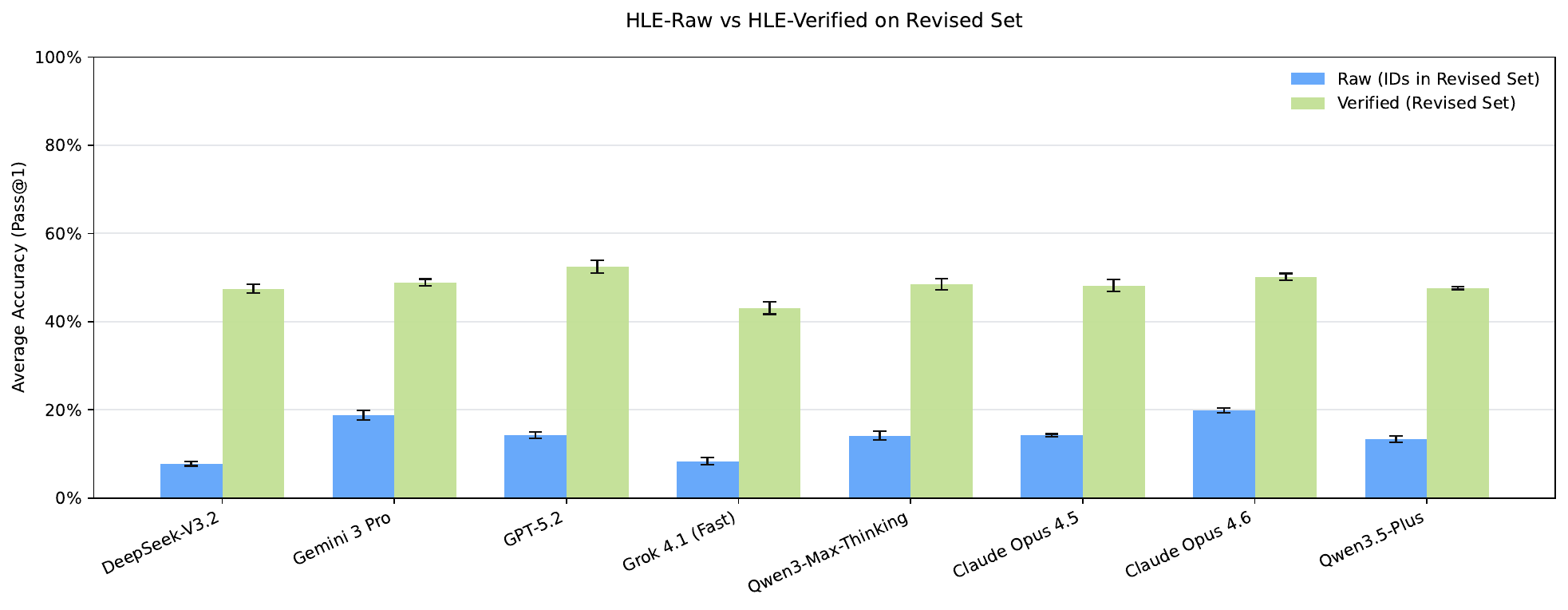}
  \caption{HLE Raw vs HLE-Verified on Revised Set}
  \label{fig:leaderboard_revised_set}
\end{figure}

\begin{figure}[t]
  \centering
    \includegraphics[width=0.3\linewidth]{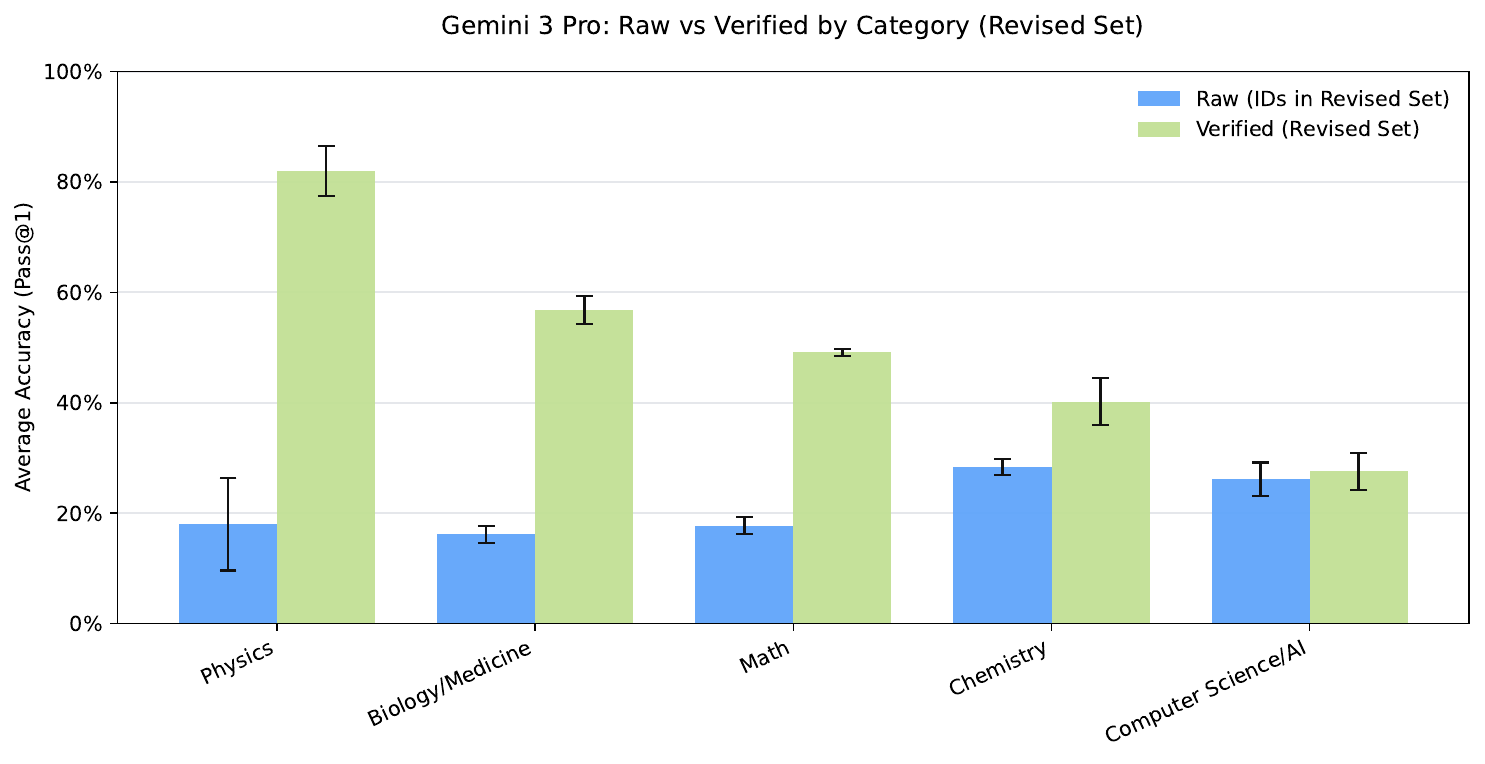}
    \includegraphics[width=0.3\linewidth]{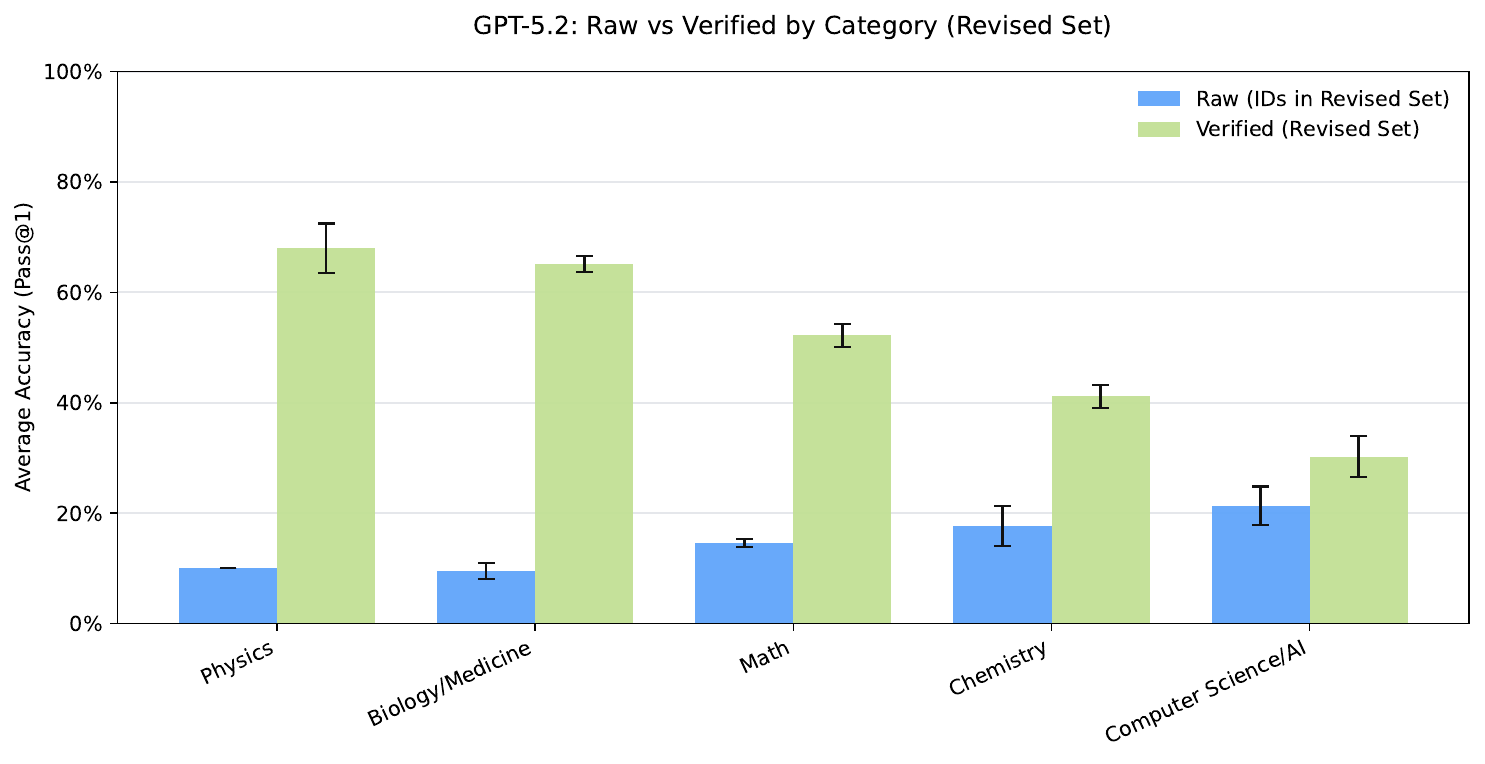}
    \includegraphics[width=0.3\linewidth]{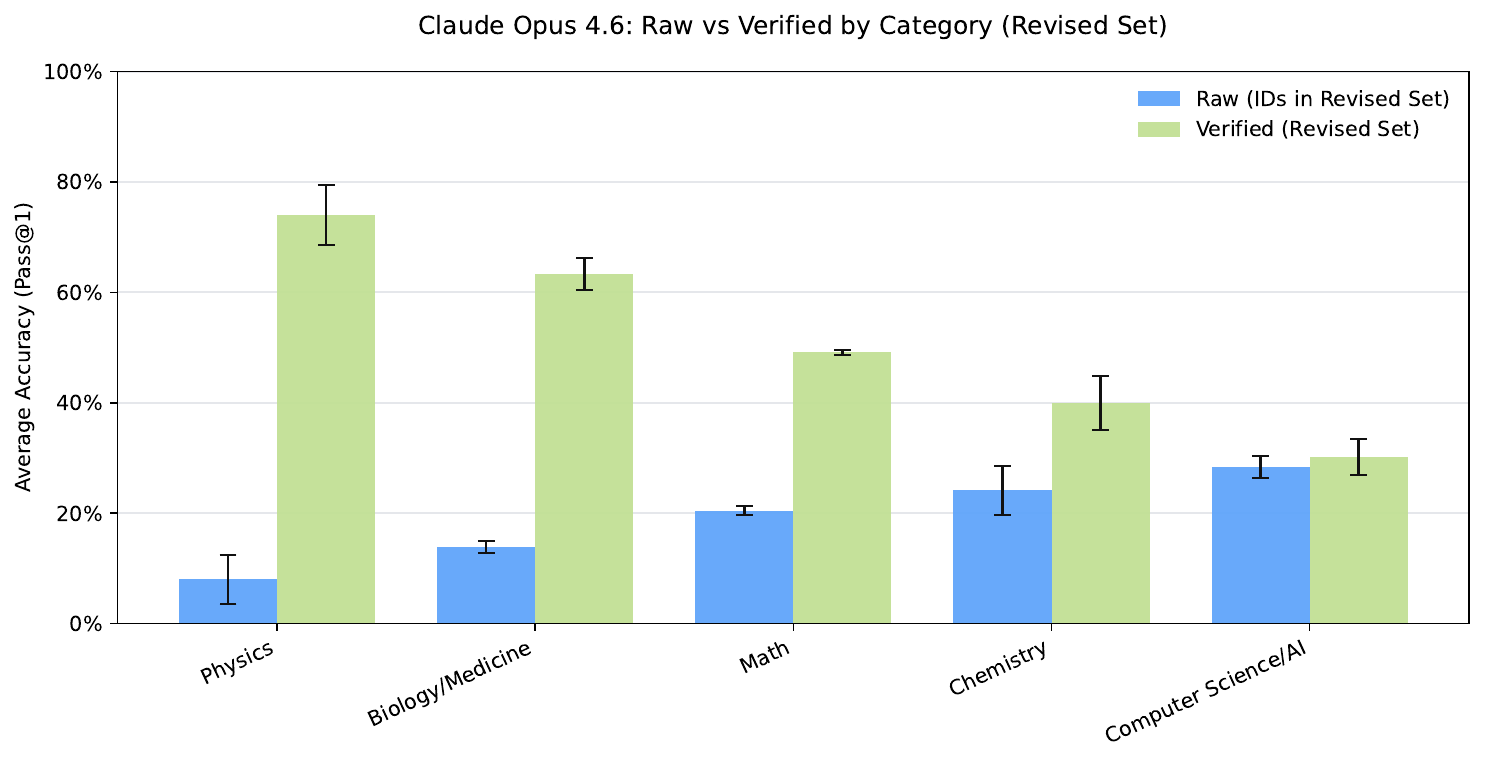}
    \includegraphics[width=0.3\linewidth]{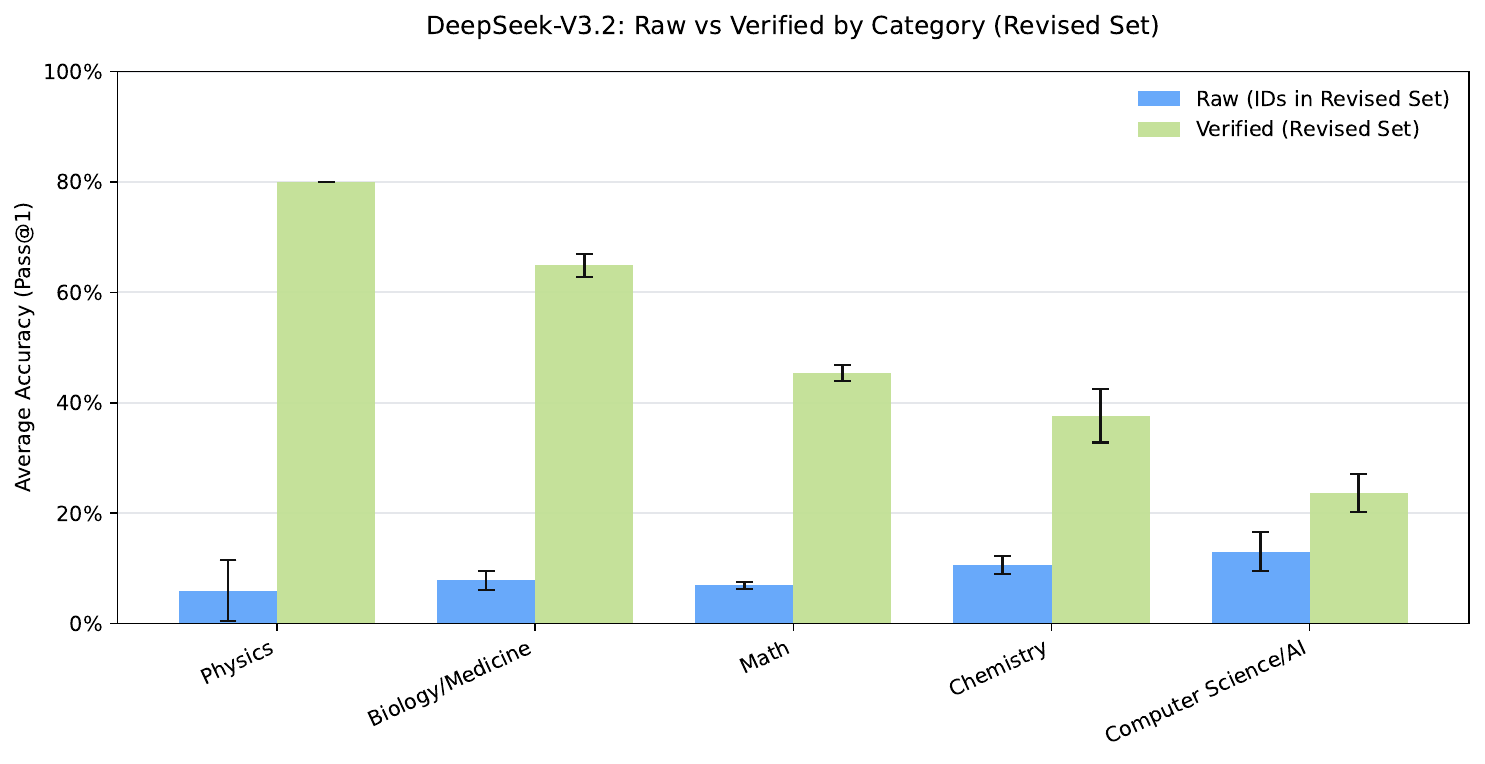}
    \includegraphics[width=0.3\linewidth]{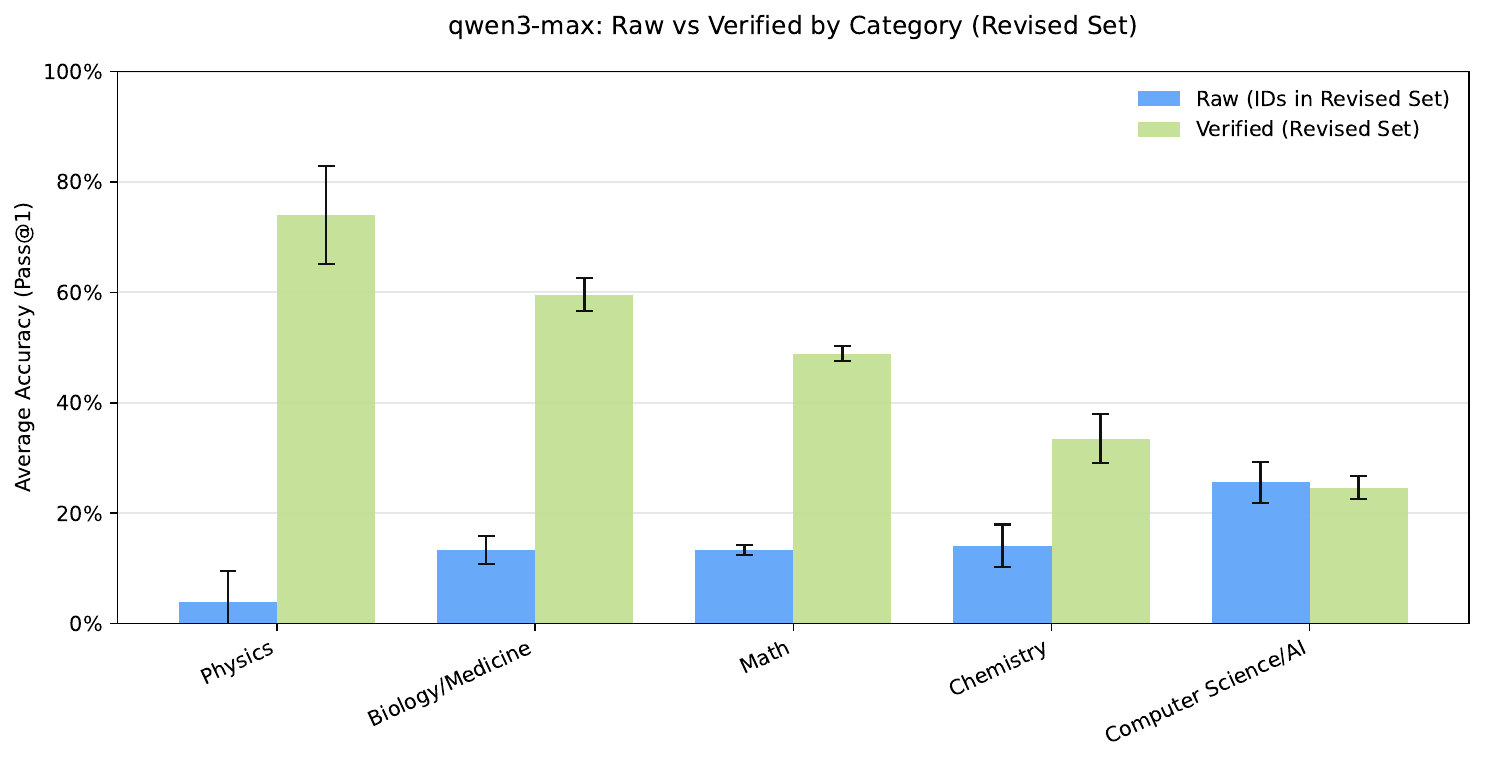}
    \includegraphics[width=0.3\linewidth]{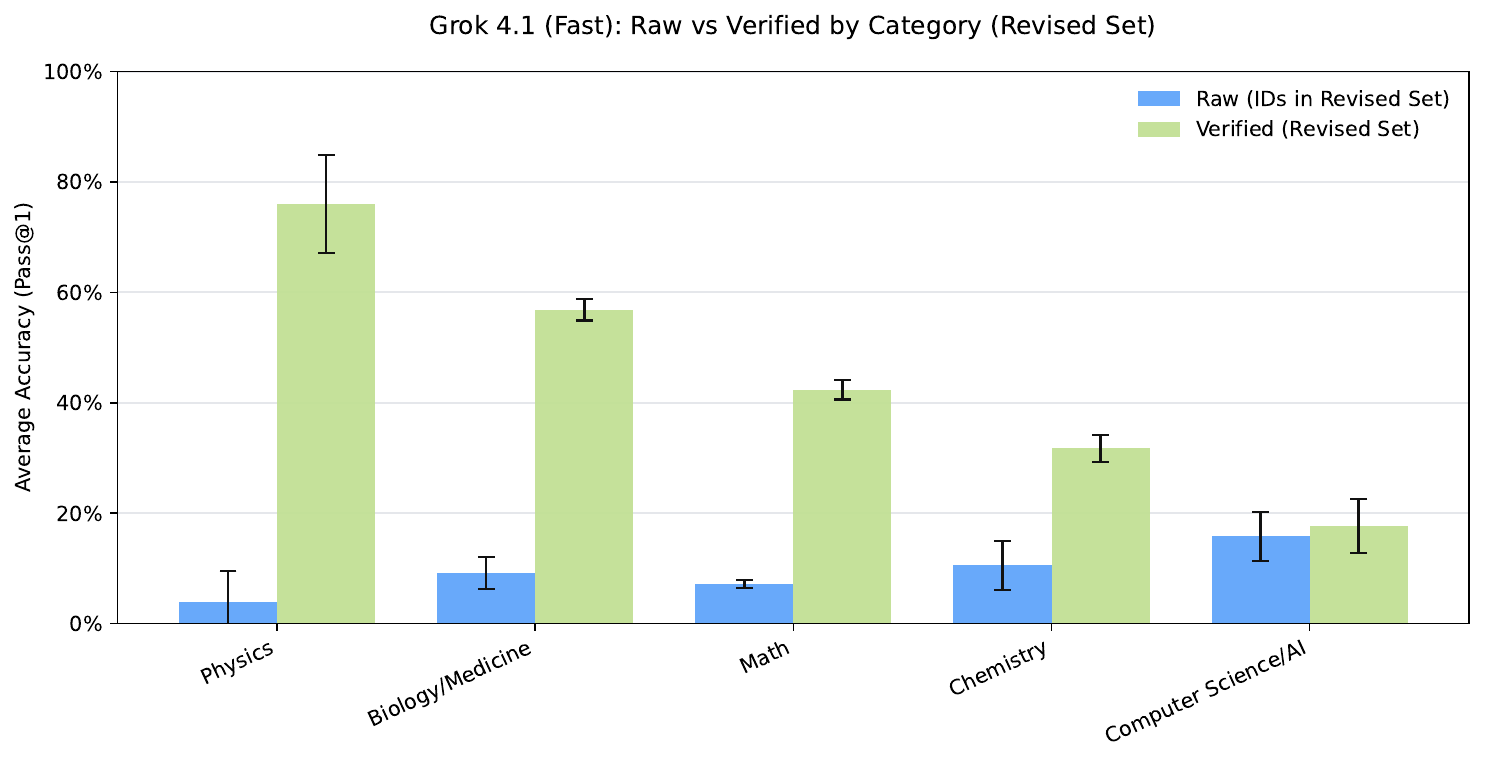}
  \caption{HLE Raw vs HLE-Verified on Revised Set across Subject Categories.}
  \label{fig:leaderboard_revised_set_category}
\end{figure}

\begin{figure}[t]
  \centering
    \includegraphics[width=0.7\linewidth]{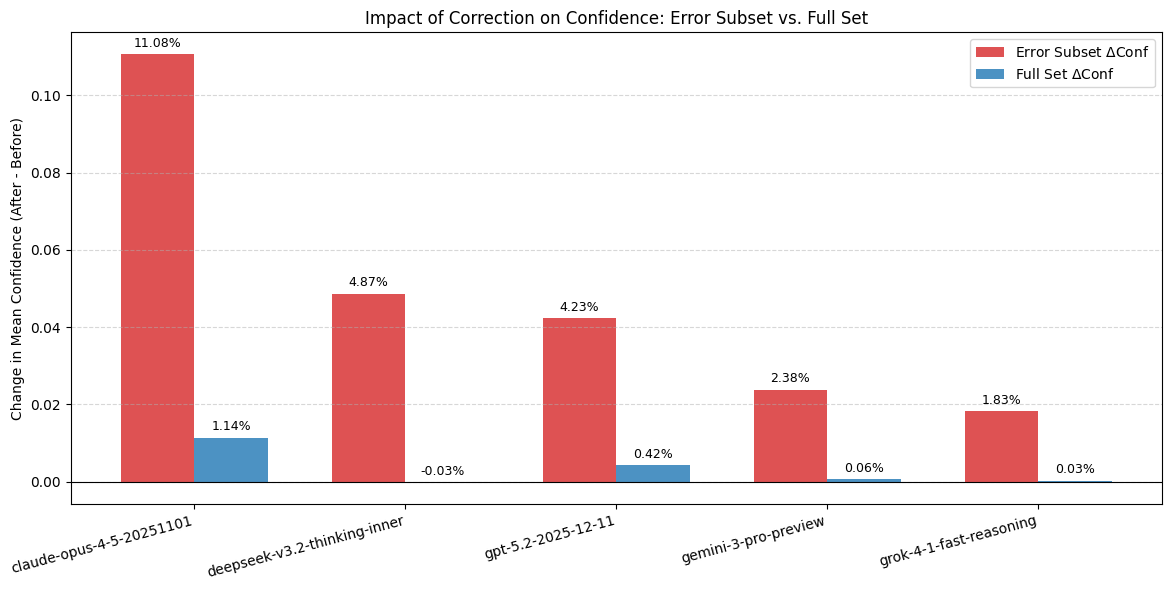}
  \caption{\textbf{Impact of item repair on model confidence.}
Mean confidence shift $\Delta\text{Conf}=\mathbb{E}[c_{\text{Verified}}-c_{\text{Raw}}]$ computed on the \textbf{Full Set} (blue) and on the \textbf{Problem-Error Subset} containing items with statement-level errors (red). Confidence increases consistently after repair on the error subset, while shifts on the full set are near zero due to dilution by unchanged items.}
  \label{fig:confidence_analysis}
\end{figure}

\subsection{Comparison on Revised Subset across Subject Categories}
\paragraph{Category-level gains are uneven but consistently positive.}
Figure~\ref{fig:leaderboard_revised_set_category} further breaks down the \textbf{Revised Subset} by subject category for several representative models. Across \emph{all} categories, we observe a consistent Raw$\rightarrow$Verified accuracy increase, confirming that the benefits of verification are not confined to a single domain. However, the magnitude of improvement varies substantially by subject: the largest jumps appear in \textbf{Physics} and \textbf{Biology/Medicine}, where raw accuracies are particularly low but rise sharply after correction, suggesting that benchmark flaws in these categories more frequently affect the problem/answer fields and thus the final pass/fail outcome. In contrast, \textbf{Chemistry} and \textbf{Computer Science/AI} show smaller yet still clear gains, indicating fewer or less severe scoring-impacting issues among the revised items. Overall, this category-wise analysis highlights that raw HLE introduces \emph{non-uniform} measurement noise across subjects, and HLE-Verified mitigates this distortion, yielding a more faithful cross-domain comparison of model capability. 


\subsection{Confidence as a Diagnostic for Noisy Items}
A natural question is whether models' self-reported confidence is sensitive to noise in the benchmark itself. If noisy items systematically elicit lower confidence, confidence (and related calibration signals) may serve as a practical diagnostic for item flaws; conversely, if models remain highly confident on flawed items, confidence-based analyses may be unreliable. We therefore study the relationship between model confidence and item quality by comparing confidence statistics before and after verification and repair. To align with this goal, we focus on items whose \emph{problem statements} are flagged as erroneous, and we compare each model's confidence on the original version versus the repaired version of the \emph{same} item. Intuitively, if an item is flawed, we expect models to express lower confidence on the raw statement due to missing conditions, ambiguities, or contradictions; after repair, the clarified statement should reduce uncertainty and lead to higher confidence.

Accordingly, for each model we compute the mean confidence shift
\[
\Delta \text{Conf} \;=\; \mathbb{E}\!\left[c_{\text{Verified}} - c_{\text{Raw}}\right],
\]
reported both on the \textbf{Full Set} and on a \textbf{Problem-Error Subset} consisting only of items with statement-level errors. Figure~\ref{fig:confidence_analysis} shows that confidence increases consistently after repair on the Problem-Error Subset across all evaluated models, with gains ranging from roughly +1.83 to +11.08 confidence points (absolute). In contrast, the Full-Set shifts are near zero (and can even be slightly negative), which is expected because the majority of items are unchanged and thus dilute the effect. 

These results suggest that statement-level noise in raw HLE may not only depress accuracy but also \emph{reduce model confidence} in a systematic way. As a result, confidence (and related calibration signals) could be a useful diagnostic for flagging potentially noisy or ambiguous items, and HLE-Verified appears to mitigate one source of uncertainty that might otherwise distort confidence-based evaluation.

\section{Conclusion}
We introduced \textbf{HLE-Verified}, a verified and revised benchmark intended to strengthen the scientific reliability of HLE-based evaluations. Our work makes benchmark flaws measurable, provides a transparent correction pipeline, and quantifies how such flaws bias reported accuracy and calibration.

Beyond this release, the disputed set provides a roadmap for community-driven improvements. We expect that a continuously maintained verification process, with structured metadata and clear contribution guidelines, can make exam-style benchmarks more robust and more informative for tracking real progress in language model reasoning.

\bibliographystyle{plainnat}
\bibliography{reference}

\appendix

\setcounter{figure}{0}
\setcounter{table}{0}

\renewcommand{\thefigure}{S\arabic{figure}}
\renewcommand{\thetable}{S\arabic{table}}

\section{Appendix}
\label{app:algo}
\subsection{LLM Judge Prompt}
\subsubsection{LLM Judge Prompt in Stage I}
\begin{tcolorbox}[title=Stage I: Model Replication and Answer Verification Prompts]

\textbf{(1) Solver Prompt.}

\begin{lstlisting}[style=promptstyle]
Please solve the following problem and place the final answer inside \boxed{}.
If there are multiple sub-questions, number them and include all answers inside \boxed{}.

{question}

Provide detailed reasoning steps and mark the final answer using \boxed{}.
\end{lstlisting}

\textbf{Purpose.}  
Standardizes solver outputs and enforces structured answer formatting to facilitate downstream extraction.

\vspace{0.5em}

\textbf{(2) Answer Extraction Prompt.}

\begin{lstlisting}[style=promptstyle]
Extract the final answer from the following mathematical solution:

{response}

Requirements:
1. Extract the expression inside \boxed{} if present.
2. If no \boxed{}, extract the final numeric or declarative conclusion.
3. Remove formatting markers such as \boxed{}, $$, \(\).
4. If multiple answers exist, separate them with commas.
5. Preserve mathematical symbols such as \sqrt{}, \frac{}.

Return only the clean mathematical answer.
Answer:
\end{lstlisting}

\textbf{Purpose.}  
Separates reasoning from final answer to reduce evaluation noise.

\vspace{0.5em}

\textbf{(3) Answer Equivalence Judgment Prompt.}

\begin{lstlisting}[style=promptstyle]
Compare the following two answers for mathematical equivalence.

Problem: {question_text}
Reference Answer: {standard_answer}
Model Answer: {model_answer}

Comparison Rules:
1. Determine mathematical equivalence.
2. Consider alternative forms (e.g., 2\sqrt{3} vs 2*sqrt(3)).
3. Consider answer ordering for multi-part responses.
4. Ignore formatting differences.

Respond with exactly one:
- "Correct"
- "Incorrect"
- "Uncertain"
\end{lstlisting}

\textbf{Role in Pipeline.}  
Used to compute pass@k replication statistics.  
Outputs are treated as diagnostic signals rather than definitive correctness labels.

\end{tcolorbox}

\subsubsection{LLM Judge Prompt in Stage II}
\begin{tcolorbox}[title=Stage II: Structured Repair and Multi-Model Adjudication Prompts]

\textbf{(1) Structured Extraction from Historical Responses.}

\begin{lstlisting}[style=promptstyle]
You are a strict information extractor.

From the following solution text, extract:
1) Final answer (preferably from \boxed{}).
2) Core reasoning steps (retain key logical steps only).

If the final answer cannot be reliably extracted,
set "empty_answer" to true.

Output JSON only:
{
  "empty_answer": true/false,
  "answer": "...",
  "rationale": "..."
}

Solution text:
{full_text}
\end{lstlisting}

\vspace{0.5em}

\textbf{(2) Repair Prompt (Expert Correction Model).}

\begin{lstlisting}[style=promptstyle]
You are a rigorous problem-repair expert.

Review:
- Problem statement
- Standard answer (if any)
- Standard rationale (if any)
- Historical model responses

Your task is to output the correct answer and reasoning.

Risk Control:
- If uncertain, set "empty_answer" to true.
- Do not guess.

Output JSON:
{
  "empty_answer": true/false,
  "answer": "...",
  "rationale": "...",
  "confidence": 0.0-1.0,
  "notes": "Brief explanation"
}
\end{lstlisting}

\vspace{0.5em}

\textbf{(3) Final Adjudication Prompt.}

\begin{lstlisting}[style=promptstyle]
You are the final adjudicator.

Choose among A/B/C repair candidates,
or choose "EMPTY" if none are reliable.

Criteria:
- Self-consistent
- Verifiable
- Matches problem statement
- Prefer EMPTY over fabrication

Output JSON:
{
  "final_choice": "A"|"B"|"C"|"EMPTY",
  "empty_answer": true/false,
  "answer": "...",
  "rationale": "...",
  "reason": "One-sentence justification"
}
\end{lstlisting}

\textbf{Purpose.}  
Implements multi-model consensus and conservative arbitration to minimize false corrections.

\end{tcolorbox}
\begin{tcolorbox}[title=Stage II: Post-Repair Cross-Audit Prompt]

\begin{lstlisting}[style=promptstyle]
You are a strict quality auditor.

Evaluate the repaired item:

(1) question_correct:
Is the problem statement clear, self-consistent, and sufficient?

(2) answer_correct:
Is the repaired answer correct and consistent with the problem?

(3) rationale_correct:
Does the repaired reasoning logically derive the answer?

If uncertain, choose false conservatively.

Output JSON only:
{
  "question_correct": true/false,
  "answer_correct": true/false,
  "rationale_correct": true/false,
  "reason_question": "...",
  "reason_answer": "...",
  "reason_rationale": "..."
}
\end{lstlisting}

\textbf{Role in Pipeline.}  
Provides structured post-repair validation and closes the verification loop under a conservative decision policy.

\end{tcolorbox}

\subsection{Quality control and decision principles}

We adopt conservative decision principles throughout both stages, reflecting the asymmetric risk inherent in benchmark release. Including a flawed item can introduce systematic evaluation bias and distort cross-model comparisons, whereas excluding a potentially valid item primarily reduces coverage without inducing measurement error. Consequently, our protocol prioritizes minimizing false inclusion over maximizing dataset size.

\paragraph{Conservative inclusion.}
Inclusion into the gold or revision subsets requires positive evidence of component-wise validity. Items are admitted only when the \textbf{problem} and \textbf{final answer} are judged well-posed, correct, and stable under expert scrutiny. The absence of detected defects is not considered sufficient; rather, validity must be affirmatively supported. Rationale-level defects alone do not automatically disqualify an item, since evaluation is typically answer-based. However, when a rationale defect signals deeper ambiguity, hidden assumptions, or incompatibility with the final answer, the item is escalated for re-audit or routed to revision.

\paragraph{Model-assisted checks as auxiliary evidence.}
Model-based replication outcomes (e.g., pass@8 success rates) are treated as diagnostic signals rather than adjudicative authority. Extreme patterns—such as systematic failure across multiple strong solvers—may trigger expert re-audit for underspecification, hidden conventions, or answer-key errors. Conversely, high solver agreement is not regarded as proof of correctness. Models serve as stochastic probes that expose potential instability, but final correctness judgments remain grounded in expert evaluation.

\paragraph{Expert recruitment and profile.}
The verification and revision process was conducted by domain experts recruited through two independent supplier teams and supported by internal adjudication specialists. The participating experts are predominantly Master's- and Ph.D.-level researchers from research-intensive universities and institutes, with formal academic training in mathematics, physics, chemistry, biomedicine, and computer science. Their academic backgrounds align directly with the disciplinary scope of HLE-Verified, enabling subject-matter–grounded validation and revision across all covered domains. Independent acceptance reviewers with doctoral-level expertise further examined complex or contentious cases.

\subsection{Case Study}
\bigskip





\begin{tcolorbox}[title=Supplementary Case 1: Constraint Violation in BNLJ Cost Modeling (Database Systems)]
\textbf{Defect Tags:} S7 Missing Prerequisite + A1 Incorrect Answer

\textbf{Error Mechanism.}  
The solution reduced page counts using selection predicates while the query explicitly required execution \emph{without materialization}.  
This omitted constraint altered the computational model.

\textbf{Revision.}  
Applying the canonical BNLJ formula:
\[
\text{Cost} = N_R + \left\lceil \frac{N_R}{B-2} \right\rceil N_S,
\]
yields the correct minimum of 465 I/Os.

\textbf{Significance.}  
Demonstrates how omission of a single prerequisite invalidates answer correctness.
\end{tcolorbox}









\begin{tcolorbox}[title=Supplementary Case 2: Structural Invariant Violation (Mathematics)]
\textbf{Defect Tags:} A1 Incorrect Answer

\textbf{Error Mechanism.}  
A closed-form “standard answer” contradicted invariants implied by the Euler sequence.  
Enumeration for small $n$ exposed systematic divergence.

\textbf{Revision.}
\[
\dim H^0\!\left(\mathbb{P}^n,\Omega^1_{\mathbb{P}^n}\otimes\mathcal{O}(2)\right)
=\frac{n(n+1)}{2}.
\]

\textbf{Significance.}  
Illustrates that symbolic plausibility does not guarantee theoretical correctness; invariant validation is essential.
\end{tcolorbox}

\begin{tcolorbox}[title=Supplementary Case 3: Anatomical Mislocalization (Biology/Medicine)]
\textbf{Defect Tags:} S3 Empirical Soundness Violation + A1 Incorrect Answer

\textbf{Error Mechanism.}  
A complete oculomotor palsy case was incorrectly localized to the reticular formation, contradicting established neuroanatomy.

\textbf{Revision.}  
Correct localization: \textbf{midbrain}, containing the oculomotor nuclear complex.

\textbf{Significance.}  
Restores clinicopathological coherence and prevents domain-level misinformation in medical evaluation.
\end{tcolorbox}

\bigskip

\subsection{Component-wise Defect Distribution Across Subjects}
\begin{figure}[htbp]
  \centering
  \includegraphics[width=0.98\linewidth]{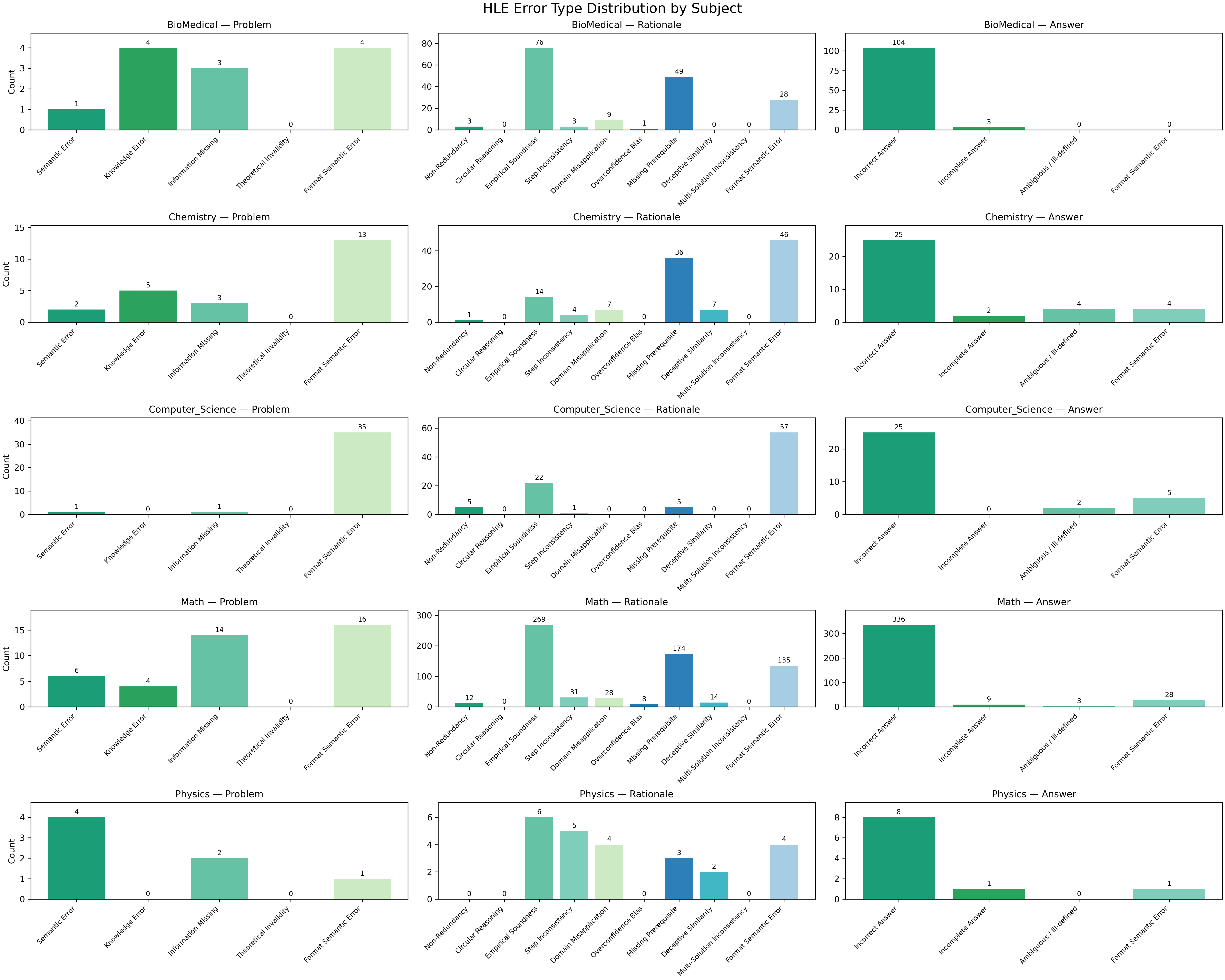}
  \caption{\textbf{HLE Error Type Distribution by Subject}}
  \label{fig:error_dist_subjects}
\end{figure}

\subsection{Cross-subject proportional differences.}

Figure~\ref{fig:error_dist_subjects} reveals substantial domain-level variation.

\textbf{Mathematics.}
Answer-level defects are overwhelmingly dominated by \emph{Incorrect Answer} (89.4\%). At the rationale level, type 3 (40.1\%) is the leading defect, reflecting structural incompleteness in derivations. Problem-level defects are more moderate in scale, with type 5 (40.0\%) most frequent. This pattern suggests that mathematical items are generally conceptually stable but vulnerable to answer-key inaccuracies and incomplete reasoning articulation.

\textbf{Physics.}
Physics exhibits relatively small defect counts overall. Answer-level errors remain dominated by type 1 (80.0\%). Problem-level defects are led by type 1 (57.1\%), while rationale defects show a flatter distribution with type 3 accounting for 25.0\%. Compared to other domains, physics demonstrates comparatively less concentration in a single rationale category.

\textbf{Chemistry.}
Chemistry shows a more balanced defect structure. Although incorrect answers (71.4\%) remain the leading answer-level issue, both problem- (56.5\%) and rationale-level (40.0\%) defects are dominated by format semantic errors. This reflects the domain’s sensitivity to symbolic precision and representational alignment.

\textbf{Biology/Medicine.}
Biology/Medicine displays the highest dominance of incorrect answers (97.2\%). Rationale-level defects are primarily type 3 (45.0\%), indicating incomplete explanatory structure. Problem-level defects are more evenly distributed, with type 2 (33.3\%) emerging as the most frequent category.

\textbf{Computer Science.}
Computer Science exhibits the most pronounced representation sensitivity. Format semantic errors (type 5) dominate problem-level defects (94.6\%), while type 10 dominates rationale-level defects (63.3\%). Answer-level incorrectness (78.1\%) remains substantial but less extreme than in mathematics or biomedicine.

\end{document}